%% file: main.tex
\pdfoutput=1
\documentclass[10pt,twocolumn,letterpaper]{article}

\usepackage[accsupp]{axessibility}  
\usepackage[pdftex]{graphicx}
\usepackage{iccv}
\usepackage{times}
\usepackage{epsfig}
\usepackage{subfig}
\usepackage{amsmath}
\usepackage{amssymb}
\usepackage{float}
\usepackage{multirow}
\usepackage{makecell}
\usepackage{tabularx}
\usepackage{booktabs}
\usepackage{amsfonts}       
\usepackage{nicefrac}       
\usepackage{microtype}
\usepackage{tabularray}
\usepackage{amsthm}

\newcommand{\rulesep}{\unskip\ \vrule\ }

\usepackage[pagebackref=true,breaklinks=true,letterpaper=true,colorlinks,bookmarks=false]{hyperref}
\iccvfinalcopy 

\def\iccvPaperID{11186} 

\ificcvfinal\fi

\begin{document}

\title{Probabilistic Precision and Recall\\ Towards Reliable Evaluation of Generative Models}

\author{Dogyun Park\\
Department of Computer Science, Korea University\\
Seoul, South Korea\\
{\tt\small gg933@korea.ac.kr}
\and
Suhyun Kim\thanks{corresponding author}\\
Korea Institute of Science and Technology\\
Seoul, South Korea\\
{\tt\small dr.suhyun.kim@gmail.com}
}

\maketitle
\ificcvfinal\thispagestyle{empty}\fi

\input{Sections/0_Abstract.tex}
\input{Sections/1_Introduction.tex}
\input{Sections/2_Preliminary_ver0.3.tex}
\input{Sections/3_Method_ver0.2.tex}
\input{Sections/4_Experiments.tex}

{\small
\bibliographystyle{ieee_fullname}
\bibliography{main.bbl}
}

\newpage
\appendix
\onecolumn

\input{Sections/5_Appendix.tex}

\end{document}

%% file: Sections/0_Abstract.tex
\begin{abstract}
   Assessing the fidelity and diversity of the generative model is a difficult but important issue for technological advancement. So, recent papers have introduced k-Nearest Neighbor ($k$NN) based precision-recall metrics to break down the statistical distance into fidelity and diversity. While they provide an intuitive method, we thoroughly analyze these metrics and identify oversimplified assumptions and undesirable properties of kNN that result in unreliable evaluation, such as susceptibility to outliers and insensitivity to distributional changes. Thus, we propose novel metrics, P-precision and P-recall (PP\&PR), based on a probabilistic approach that address the problems. Through extensive investigations on toy experiments and state-of-the-art generative models, we show that our PP\&PR provide more reliable estimates for comparing fidelity and diversity than the existing metrics. The codes are available at \url{https://github.com/kdst-team/Probablistic_precision_recall}.
\end{abstract}

%% file: Sections/1_Introduction.tex
\section{Introduction}
\label{sec:intro}

Modern state-of-the-art generative models such as Variational Autoencoders (VAEs) \cite{KingmaW13,razavi2019generating}, Generative Adversarial Networks (GANs) \cite{goodfellow2020generative,karras2019style,karras2020training,karras2021alias}, and Score-based Diffusion Models (SDMs) \cite{ho2020denoising,kingma2021variational,song2021maximum,dhariwal2021diffusion,vahdat2021score,rombach2022high,song2021scorebased} have shown outstanding achievements that surpass human evaluative capabilities. Consequently, reliable evaluation metrics for comparing the performance of generative models are critical for further technological advancement. 
The widely used Fr$\acute{e}$chet Inception Distance (FID) evaluation metric \cite{heusel2017gans} has been praised for its robustness and consistency with human perceptual evaluation. However, FID only provides a single score and does not diagnose whether the model lacks fidelity or diversity, two essential aspects for improving generative models. Recent papers \cite{sajjadi2018assessing,simon2019revisiting,kynkaanniemi2019improved,naeem2020reliable,alaa2022faithful} have introduced various two-value metrics to measure fidelity and diversity, respectively. Especially,  $k$-nearest neighbor ($k$NN) based metrics such as Improved Precision and Recall (IP\&IR)~\cite{kynkaanniemi2019improved} and Density and Coverage (D\&C)~\cite{naeem2020reliable} have been used to assess fidelity and diversity on various generative models \cite{dhariwal2021diffusion,zheng2021spatially,anokhin2021image,razavi2019generating,karras2020training,chan2021pi,ntavelis2022arbitrary}. The pipeline of these metrics \cite{kynkaanniemi2019improved,naeem2020reliable} starts with estimating distribution support of real and fake distribution by constructing sample-specific hyperspheres around each sample using $k$NN. Then, they 
measure the expectation of the likelihood of fake (real) samples with respect to the estimated real (fake) distribution support to represent fidelity (diversity). 

\begin{figure*}[t]
\centering
\includegraphics[width=0.43\linewidth]{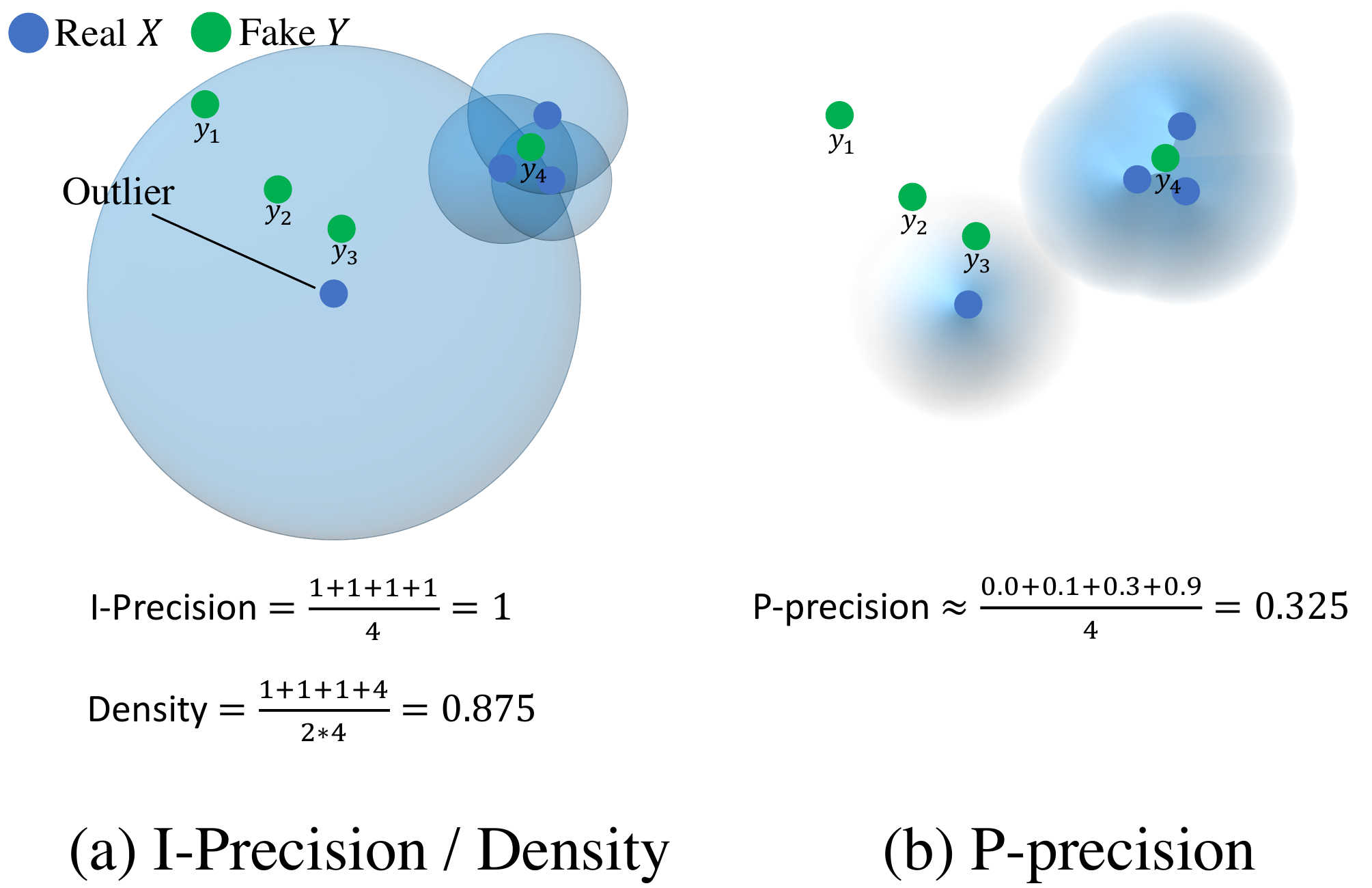}
\hspace{5pt}
\rulesep
\hspace{5pt}
\includegraphics[width=0.505\linewidth]{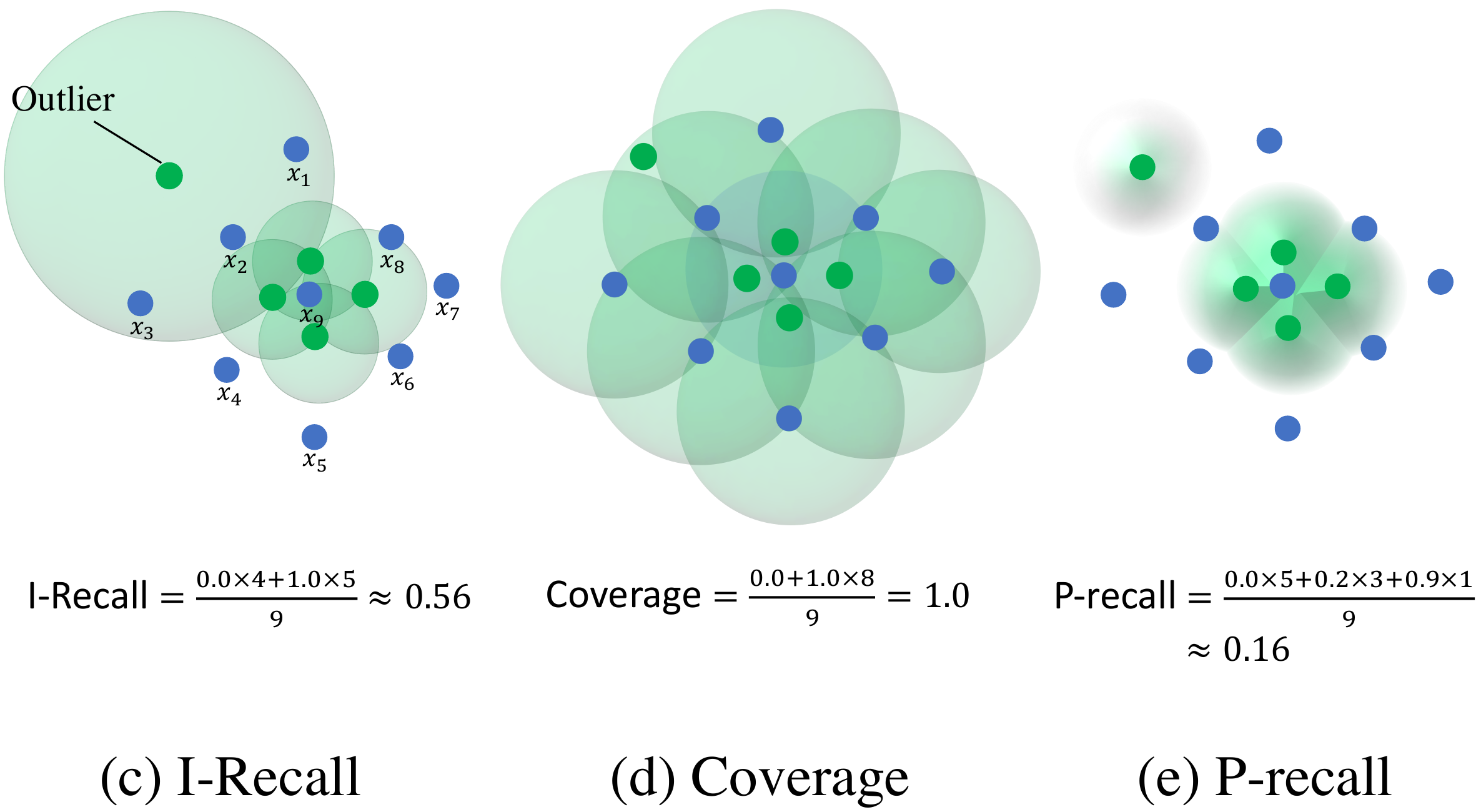}
\caption{\textbf{Examples of IP\&IR, D\&C, and PP\&PR} (better zoom-in). For simplicity, we use $k=2$ for $k$-Nearest Neighbor in all metrics. (a), (c): Due to the overestimation of $k$NN by the outlier and constant-density assumption within hyperspheres, I-precision, I-recall, and Density denote the same values for different $y$ in an overly large space (See $y_1$, $y_2$, and $y_3$ in (a)), leading to unreliably-overestimated values. (b), (e): On the other hand, our P-precision and P-recall assign different scores to different $y$ based on a probabilistic approach and address overestimation of $k$NN (See Sec.~\ref{sec:method} for details). (d): This illustrates a case when Coverage exhibits its conceptual limitation. Even if the fake samples have lower relative diversity compared to real samples, they can still be included in multiple hyperspheres of real samples, leading to high Coverage.}
\label{fig:main1}
\end{figure*}

Unfortunately, both IP\&IR and D\&C metrics present certain challenges that compromise their practical reliability. Specifically, IP\&IR operate under the assumption of constant-density within the distribution support and overlook the overestimation linked to $k$NN, leading to an inflated distribution support with constant density. This scenario becomes especially problematic when the distribution houses outliers, resulting in metric overestimation (See Fig.~\ref{fig:main1}a).
Density metric \cite{naeem2020reliable} attempts to address these limitations by accumulating density over overlapped $k$NN hyperspheres. Yet, our findings suggest that it still remains unreliable to outliers. Furthermore, it tends to exhibit high variance, due to its extremely wide bound, which further limits the reliability of evaluations.
On the other hand, Coverage metric \cite{naeem2020reliable} presents an alternative method that measures robust diversity even with the presence of outliers in the fake distribution. However, we show that it suffers from a conceptual limitation that makes it insensitive to changes in fake distribution. An in-depth discussion of these constraints is provided in Sec.~\ref{sec:background}.
In the end, these issues underscores the need for careful revision of $k$NN-based metrics to provide more reliable evaluations.

We, therefore, propose a novel set of metrics, P-precision and P-recall (PP\&PR), which aim to provide more reliable estimations for the fidelity and diversity of generative models.
In particular, we derive our metrics based on a \textit{probabilistic approach} that relaxes the oversimplified assumptions and undesirable properties of $k$NN, addressing the limitations of existing metrics.
Through extensive experiments on both toy and real-world datasets, we systematically demonstrate the superiority of our proposed metrics over IP\&IR and D\&C. To summarize, our contributions are as follows:
\begin{itemize}
    \item We propose a set of two-value metrics, P-precision and P-recall, that provide reliable fidelity and diversity measures for generative models.
    \item We thoroughly analyze the limitations of existing metrics (IP\&IR and D\&C), such as vulnerability to outliers and insensitivity to diversity, and introduce a novel method based on the probabilistic approach that remedies the limitations. 
    \item Through a series of experiments on toy and real-world datasets with state-of-the-art generative models, we empirically verify the limitations of current metrics and demonstrate the efficacy of our proposed metrics.
\end{itemize}

%% file: Sections/2_Preliminary_ver0.3.tex
\section{Preliminary}\label{sec:background}
\subsection{Feature embedding for evaluation}\label{section:pipeline}
Statistical comparison of high dimensional data in the raw input space $\mathcal{X}$ is difficult. Therefore, the usual evaluation starts with converting real data $X\sim P$ and fake data $Y\sim Q$ into embeddings using a CNN feature $\Phi$ $(\Phi(X)$ and $\Phi(Y))$ \cite{heusel2017gans,kynkaanniemi2019improved,salimans2016improved,naeem2020reliable}. This embedded feature space is believed to have more ``meaningful" information to compare two samples \cite{forsyth2012computer,xu2018empirical}. 
For simplicity, we use $X$ and $Y$ in place of $\Phi(X)$ and $\Phi(Y)$ for the remainder of the paper.

\subsection{$k$NN-based fidelity and diversity measures}
Now, we briefly introduce current $k$NN-based fidelity and diversity measures and discuss their limitations in Sec.~\ref{sec:background_problems}.

\paragraph{Improved Precision and Recall (IP\&IR).}
To represent fidelity, Precision is defined as the expected likelihood of a fake sample belonging to the support of the real distribution, which refers to the set of possible real samples whose probability density function is non-zero. Likewise, Recall is defined as the expected likelihood of a real sample belonging to the support of fake distribution to measure diversity.
\begin{equation}\label{eq:main6}
\begin{split}
    \text{Precision} &:= \mathbb{E}_{Y\sim Q}\left[Y \in \text{supp}(P)\right] \\
    \text{Recall} &:= \mathbb{E}_{X\sim P}\left[X \in \text{supp}(Q)\right], 
\end{split}
\end{equation} 
where $\text{supp}(\cdot)$ indicates the ground truth support of distribution. Then, IP\&IR~\cite{kynkaanniemi2019improved} approximates them using the empirical observations $\left(\{x_i\}_{i=1}^N \ \text{and} \ \{y_j\}_{j=1}^M\right)$ as  
\begin{equation}
\begin{split}
    \text{Precision} &\approx \text{I-Precision} \\
     &:= \frac{1}{M} \sum_{j=1}^M \text{BSR}_P(y_j) \\
    \text{Recall} &\approx \text{I-Recall} \\ 
     &:= \frac{1}{N} \sum_{i=1}^N \text{BSR}_Q(x_i).
\end{split}
\end{equation} 
The scoring rule of IP\&IR, which we name as \textbf{B}inary \textbf{S}coring \textbf{R}ule (BSR) for comparison with other scoring rules, is
\begin{equation}\label{eq:main1}
    \text{BSR}_P(y_j) =
    \begin{cases}
    1, & \text{if} \ y_j \in \cup_{i=1}^N B\left( x_i, \text{NND}_k(x_i) \right) \\
    0, & \text{otherwise}
    \end{cases}
\end{equation}
where $B(x,r)$ is the hypersphere in $\mathbb{R}^D$ around $x$ with radius $r$, and $\text{NND}_k(x_i)$ denotes the Euclidean distance from $x_i$ to the $k^{th}$ nearest neighbor among $\{x_i\}_{i=1}^{N}$ excluding itself. In other words, $\text{BSR}_P(y_j)$ assigns a binary score of 1 or 0 to $y_j$ based on whether it falls inside the union of hyperspheres.
It is worth noting that BSR employs a \textit{deterministic approach} that determines the support as the union of $k$NN hyperspheres without accounting for any uncertainty regarding the true support. This restricts the scoring rule to binary assignments, leading to a constant density within the approximated distribution support.

I-recall is defined using $\text{BSR}_Q(x_i)$, which is defined symmetrically with $\text{BSR}_P(y_j)$. It builds hyperspheres around fake samples and tests real samples.

\paragraph{Density and Coverage (D\&C).}
D\&C use different approaches to define fidelity and diversity in comparison to IP\&IR.  
Specifically, Density \cite{naeem2020reliable} tackles the issue of constant-density over distribution support in IP\&IR by quantifying the degree of overlap between $k$NN hyperspheres. To this end, Density defines a scoring rule that estimates the density function within distribution support by the superposition of hyperspheres, as opposed to the union in IP\&IR. We denote this scoring rule as \textbf{D}ensity \textbf{S}coring \textbf{R}ule (DSR).
\begin{equation}\label{eq:main3}
\begin{split}
    \text{DSR}_P(y_j) = \frac{1}{k}\sum_{i=1}^N 1_{y_j \in B \left( x_i, \text{NND}_k(x_i) \right)}
\end{split}
\end{equation}
where $k$ is for the $k$-nearest neighbors. DSR employs a kind of Parzen window density estimation \cite{hart2000pattern} with a uniform kernel whose width is adjusted automatically, except that the integral of the kernel does not sum up to 1. Now, Density is defined as the expectation of DSR over observed samples as
\begin{equation}\label{eq:main4}
\begin{split}
    \text{Density} := \frac{1}{M}\sum_{j=1}^M \text{DSR}_P(y_j)
\end{split}
\end{equation} 

For diversity, the authors employ another scoring rule, \textbf{C}overage \textbf{S}coring \textbf{R}ule (CSR), which is not symmetric to DSR. It builds $k$NN hyperspheres around the real samples and counts the real samples whose hyperspheres are occupied by the fake samples. Coverage has demonstrated effectiveness in handling outliers in fake distribution, as the hyperspheres of real samples are not influenced by fake outliers.
\begin{equation}\label{eq:main5}
    \begin{split}\ 
        &\text{CSR}_Q(x_i) = 1_{\exists j \ \text{s.t.} \ y_j \in B \left( x_i, \text{NND}_k(x_i) \right)} \\
        &\text{Coverage} := \frac{1}{N}\sum_{i=1}^N \text{CSR}_Q(x_i)
    \end{split}
\end{equation}

\section{Limitations with scoring rules of IP\&IR and D\&C}\label{sec:background_problems}
\paragraph{Binary scoring rule.} BSR can only assign constant density within an approximated distribution support, which can hinder reliable estimates when comparing distributions with complex density functions. Moreover, the use of $k$NN in BSR makes it more susceptible to outliers as it unreliably inflates the size of the hypersphere due to the presence of such outliers. This results in overestimated value as illustrated in Fig.~\ref{fig:main1}a and ~\ref{fig:main1}c. Both I-precision and I-recall face similar issues since they symmetrically use BSR.

\paragraph{Density scoring rule.} We contend that DSR remains vulnerable to outliers as it \textit{accumulates} density over the \textit{overstimated constant-density} hyperspheres. For instance, if there are $s$ outliers, all hyperspheres of outliers can overlap due to the overestimated size of hyperspheres. This causes the overlapped density to accumulate to $s/k$ (note that DSR scales the overlapped density by a factor of $1/k$ in Eq.~\ref{eq:main3}), resulting in a rapid increase with just a few outliers.
We demonstrate this quantitatively with the experiment in Sec.~\ref{sec:experiment}. Additionally, reducing the density by a factor of $1/k$ makes the metric highly dependent on the choice of $k$, which is not desirable as we demonstrate in Fig.~\ref{fig:main3_1}. 

Lastly, DSR is bounded within [0, $N/k$] (e.g., $N/k$ = $50000/5$ = $10000$), which can result in samples with abnormally high values of DSR, compromising stable evaluations with high standard deviations as demonstrated in Fig.~\ref{fig:main2}c.

\paragraph{Coverage scoring rule.} While CSR can effectively handle outliers in fake distribution, it exhibits a conceptual limitation: a lack of sensitivity to the comparative diversity between fake and real distributions. This arises because CSR doesn't factor in distances between fake samples. Instead, its focus is solely on the count of real sample hyperspheres occupied by fake samples. This leads to a scenario where if real samples are more dispersed compared to fake ones, meaning the fake samples exhibit less relative diversity, the fake samples could still be included in multiple real hyperspheres. This, in turn, can produce a falsely elevated Coverage metric (a conceptual illustration is provided in Fig.\ref{fig:main1}d). Conversely, Coverage can be low if fake samples are more widely distributed compared to real samples. We elucidate this phenomenon quantitatively in Fig.~\ref{fig:main4}b.

%% file: Sections/3_Method_ver0.2.tex
\section{Method}\label{sec:method}
To address the aforementioned limitations, we propose a novel scoring rule based on a probabilistic approach. Our motivation arises from the recognition that accurately estimating the ground truth support is infeasible with empirical observations alone. As a result, it becomes difficult to determine whether a sample belongs to the ground truth support with certainty (0 or 1). Therefore, we adopt a probabilistic perspective and define the probability of a sample belonging to the support, taking into account the uncertainty surrounding the ground truth support. 
Then, our scoring rule can assign lower values to samples that are less likely to belong to the support such as outliers, effectively addressing outlier concerns.
We also define the scoring rules for fidelity and diversity symmetrically, similar to IP\&IR, to avoid the conceptual limitation of CSR.
Although we mainly focus on describing the details of fidelity in the paper, diversity is defined symmetrically.

\subsection{P-precision and P-recall (PP\&PR)}
Given the observations $\{x_i\}_{i=1}^N$, we assume that there are myriad possible supports of $P$, denoted as $S_P$. The collection of these supports is represented by $\mathbb{S}_P$. Subsequently, we introduce \textbf{P}robabilistic-\textbf{Precision} that characterizes precision as the probability of $y_j$ being an element of support randomly chosen from $\mathbb{S}_P$. This is articulated as
\begin{equation}\label{eq:main6_1}
\begin{split}
    \text{P-precision} &:= \mathbb{E}_{Y\sim Q, S_P \in \mathbb{S}_P}\left[Y \in S_P\right] \\
    &= \mathbb{E}_{Y\sim Q}\left[\mathbb{E}_{S_P \in \mathbb{S}_P}\left[Y \in S_P\right]\right] \\
    &= \mathbb{E}_{Y\sim Q}\left[\text{Pr}(Y \in S_P)\right]
\end{split}
\end{equation}
Then, we approximate it with given observations $\{y_j\}_{j=1}^M$ as
\begin{equation}
\begin{split}
    \text{P-precision} \approx \frac{1}{M} \sum_{j=1}^M \text{Pr}(y_j\in S_P).
\end{split}
\end{equation}
In a similar manner, \textbf{P}robabilistic-\textbf{recall} is approximated as
\begin{equation}\label{eq:main6_2}
\begin{split}
    \text{P-recall} \approx \frac{1}{N} \sum_{i=1}^N \text{Pr}(x_i\in S_Q).
\end{split}
\end{equation}

\subsection{Probabilistic scoring rule} 
To estimate $\text{Pr}(y_j \in S_P)$, we first divide $S_P$ into subsets around each observation as
\begin{equation}\label{eq:main7}
\begin{split}
    S_P := \bigcup_{i=1}^N s_P(x_i),
\end{split}
\end{equation} 
where $\text{s}_P(x_i)$, denoted as the subsupport around $x_i$, represents a subset of $S_P$ that comprises elements for which the nearest reference among $\{x_i\}_{i=1}^N$ is $x_i$. 
Then, we can further elucidate $\text{Pr}(y_j \in S_P)$ as
\begin{equation}\label{eq:main9}
\begin{split}
     \text{Pr}(y_j \in S_P) &= 1- \text{Pr}(y_j \notin S_P) \\ 
     & = 1 - \text{Pr}\left(y_j \notin \cup_{i=1}^N s_P(x_i)\right) \\ 
     & = 1 - \text{Pr}\left(y_j \in \cap_{i=1}^N s_P^\mathsf{c}( x_i)\right) \\
     & = 1- \prod_{i=1}^N \text{Pr}\left(y_j \in s_P^\mathsf{c}(x_i)\right) \\ 
     & = 1- \prod_{i=1}^N \left(1-\text{Pr}(y_j \in s_P(x_i))\right)
\end{split}
\end{equation}
using Eq.~\ref{eq:main7} and complement rule, where $s_P^\mathsf{c}(\cdot)$ denotes the complement set of $s_P(\cdot)$. The equality in the fourth line of Eq.~\ref{eq:main9} comes from assuming the independence of events $\{y_j \in s^c_P(x_i) | i=1,2,\cdots, N\}$. 

Now, we define our new scoring rule termed Probabilistic Scoring Rules (PSR) from the last equation in Eq.~\ref{eq:main9}:
\begin{equation}\label{eq:maine}
\begin{split}
    \text{PSR}_P(y_j) &:= 1- \prod_{i=1}^N \left(1-\text{Pr}(y_j \in s_P(x_i))\right) \\ 
    & = \text{Pr}(y_j \in S_P) 
\end{split}
\end{equation}
In contrast with deterministic methods such as BSR, which make certain assumptions about the definitive shape of subsupport, 
PSR estimates the probability of $y_j$ to belong to the \textit{uncertain} support by combining the probability to belong to the subsupport around each observation. 
Utilizing this probabilistic approach allows us to depart from defining the specific subsupport—a task that, in practice, can be ambiguous. Rather, we infer the likelihood of a sample being part of the subsupport based on an intuitive premise: samples in closer proximity to $x_i$ are more likely to be an element of the subsupport compared to those further away. Accordingly, we approximate the probability $\text{Pr}(y_j\in s_P(x_i))$ as
\begin{equation}\label{eq:main10}
\begin{split}
     \text{Pr}(y_j\in s_P(x_i)) \approx 
     \begin{cases}
    1 - \frac{d}{R}, & \text{if} \ d \leq R \\
    0, & \text{otherwise}
    \end{cases}
\end{split}
\end{equation}
where $d$ is $l_2$ distance between $x_i$ and $y_j$. From Eq.~\ref{eq:main10}, the probability is one when $d=0$, given that we know for certain that $x_i$ belongs to the subsupport. The probability monotonically decreases with an increase in 
$d$ until it hits the threshold distance $R$, where the probability converges to zero.
By Eq.~\ref{eq:maine} and Eq.~\ref{eq:main10}, our proposed PSR can assign varying values to samples based on their likelihood of belonging to the support. Moreover, it is bounded within [0,1] (see supplement for the proof) unlike DSR which has an extremely wide bound. Note that while other distance metrics $d$ or monotonically decreasing functions could serve this purpose, we opted to use $l_2$ distance and a linear function since it was sufficiently sensitive based on our empirical results. We leave it for future work.

\begin{figure*}[t]
\centering
\includegraphics[width=0.31\textwidth]{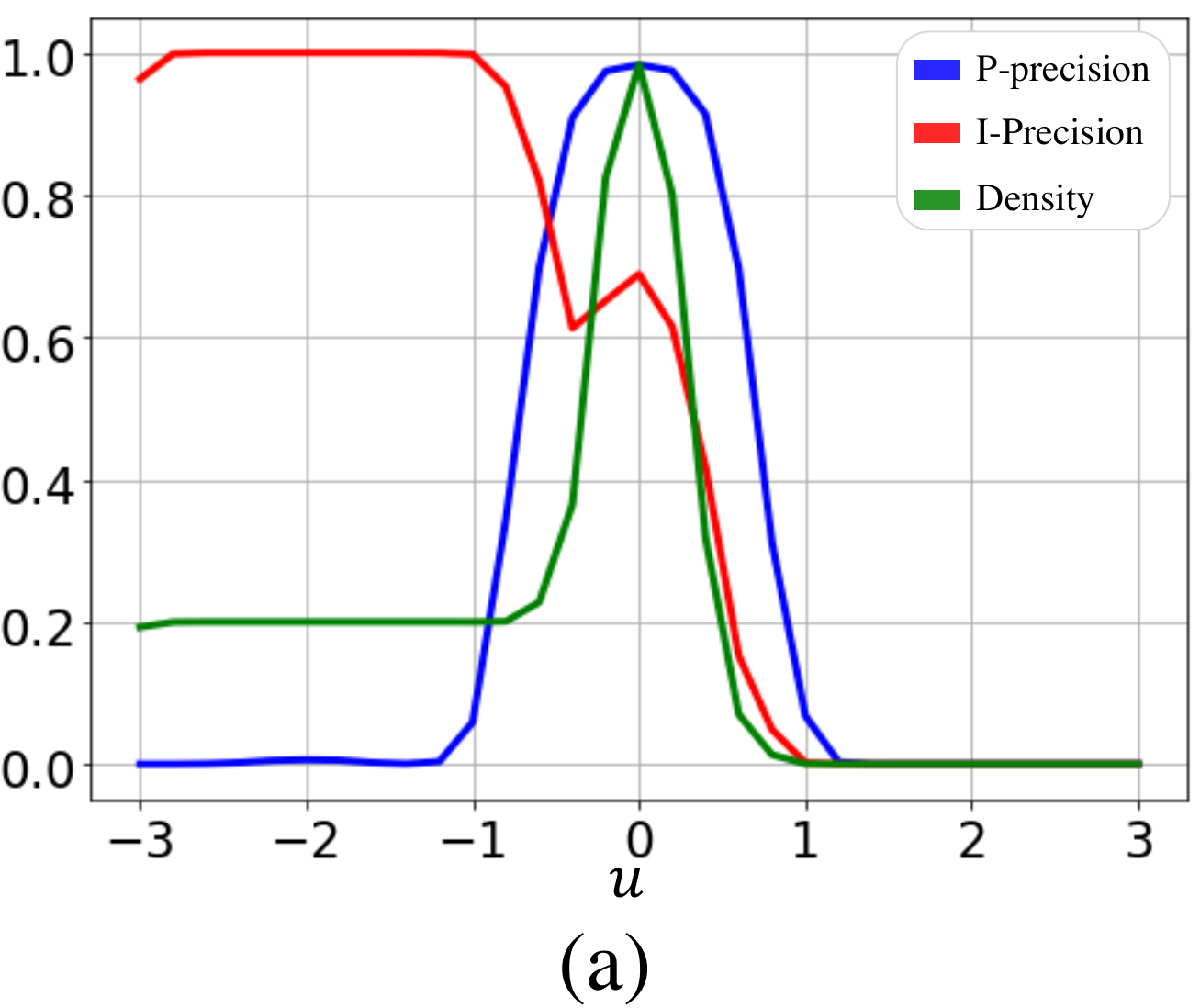} 
\includegraphics[width=0.31\textwidth]{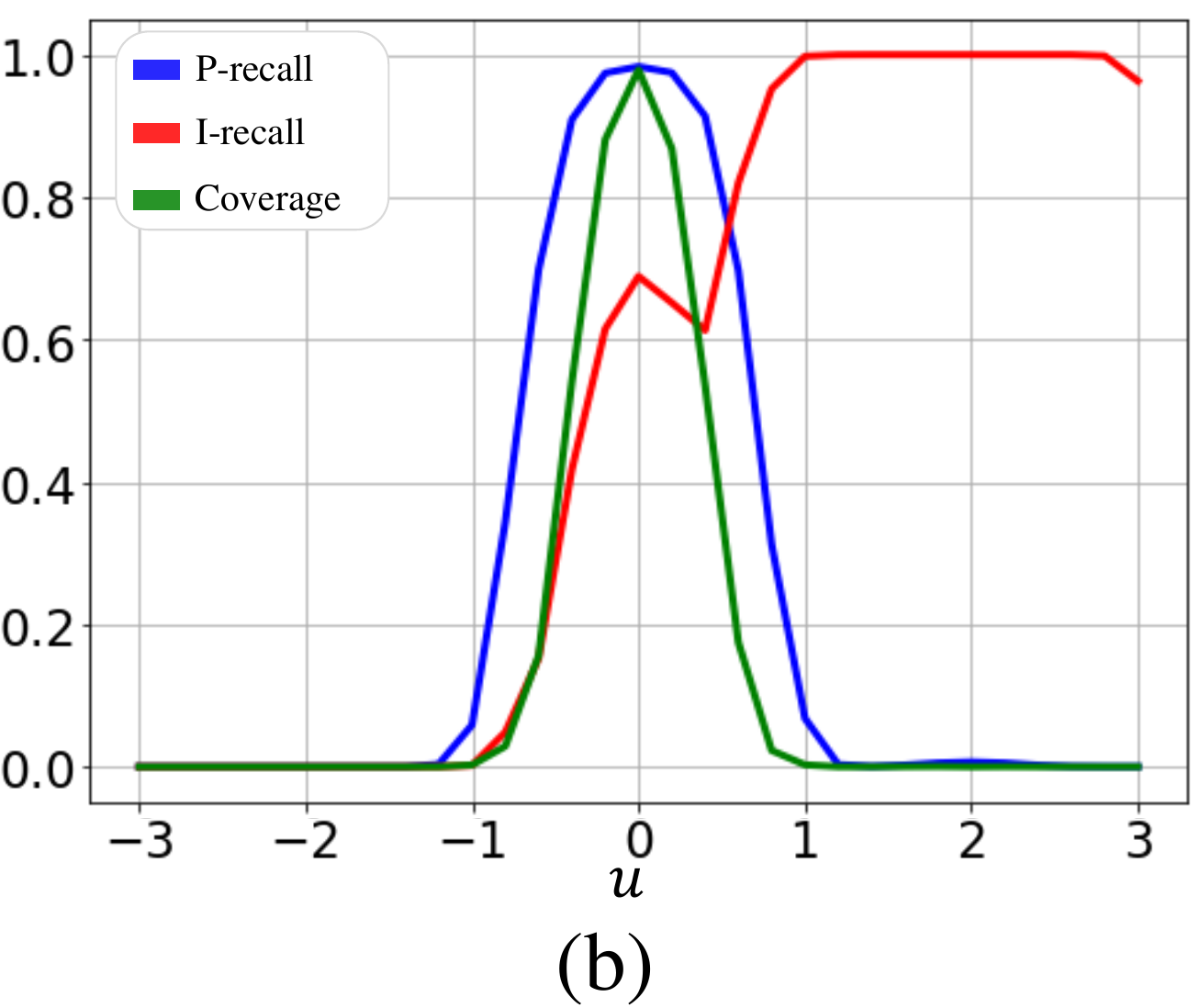}
\label{figure:outlier-real}
\includegraphics[width=0.31\textwidth]{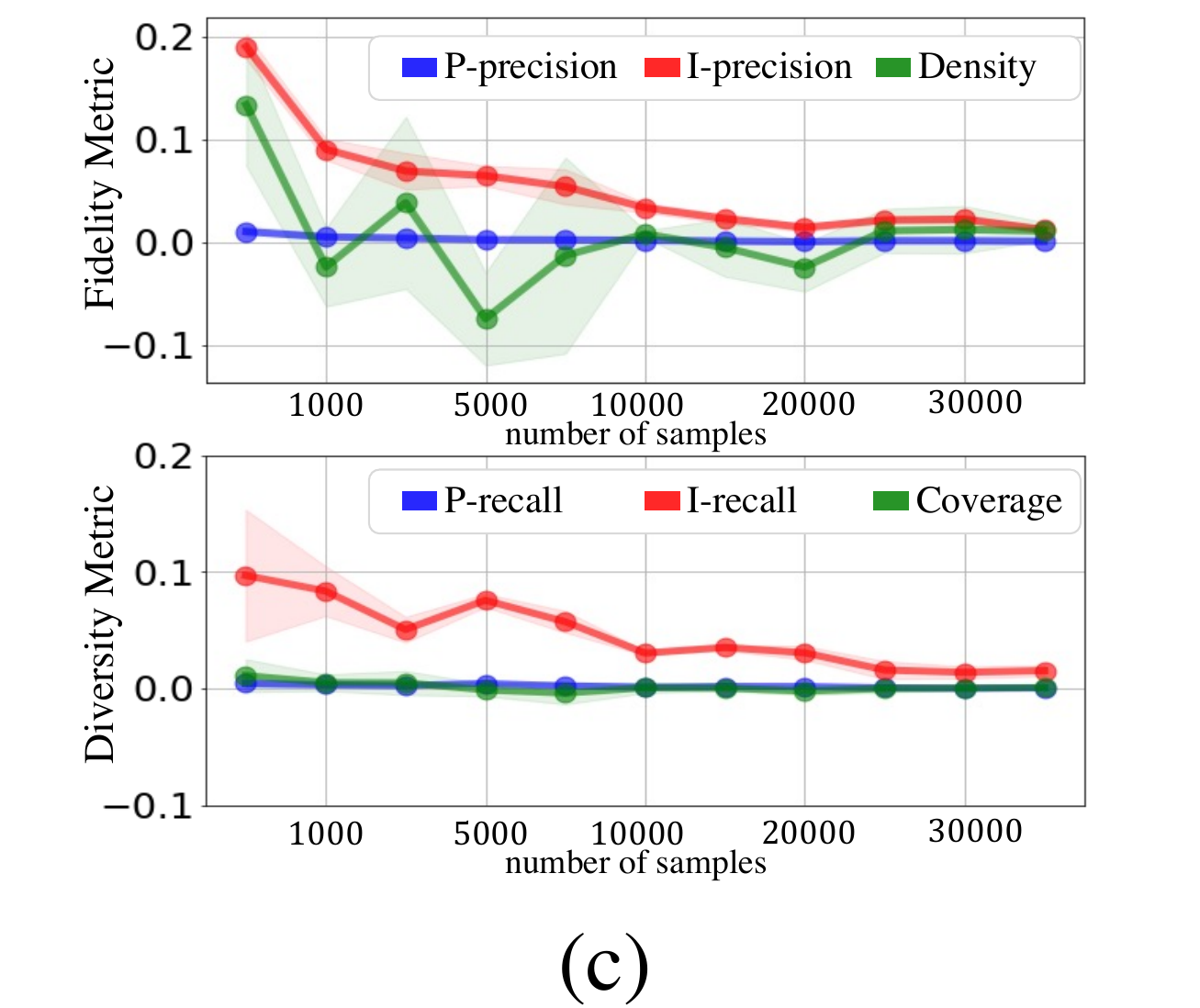}
\caption{(a) Behavior of fidelity metrics between two Gaussian distributions $X \sim N(0,I)$ and $Y \sim N(u\textbf{1},I)$ as $u$ moves between [-3,3] with outlier $x_o \sim N(-\textbf{2},I)$ added to $X$. (b) Behavior of diversity metrics between two Gaussian distributions $X \sim N(u\textbf{1},I)$ and $Y \sim N(0,I)$ as $u$ moves between [-3,3] with outlier $y_o \sim N(\textbf{2}, I)$ added to $Y$. (c) Estimated bias from the presumed true value between two identical Gaussian distributions for different numbers of datasets $N$. The line shows the means and the shaded area denotes standard deviations across 50 runs.}
\label{fig:main2}
\end{figure*}

\begin{figure}[t]
\centering
\includegraphics[width=0.95\columnwidth]{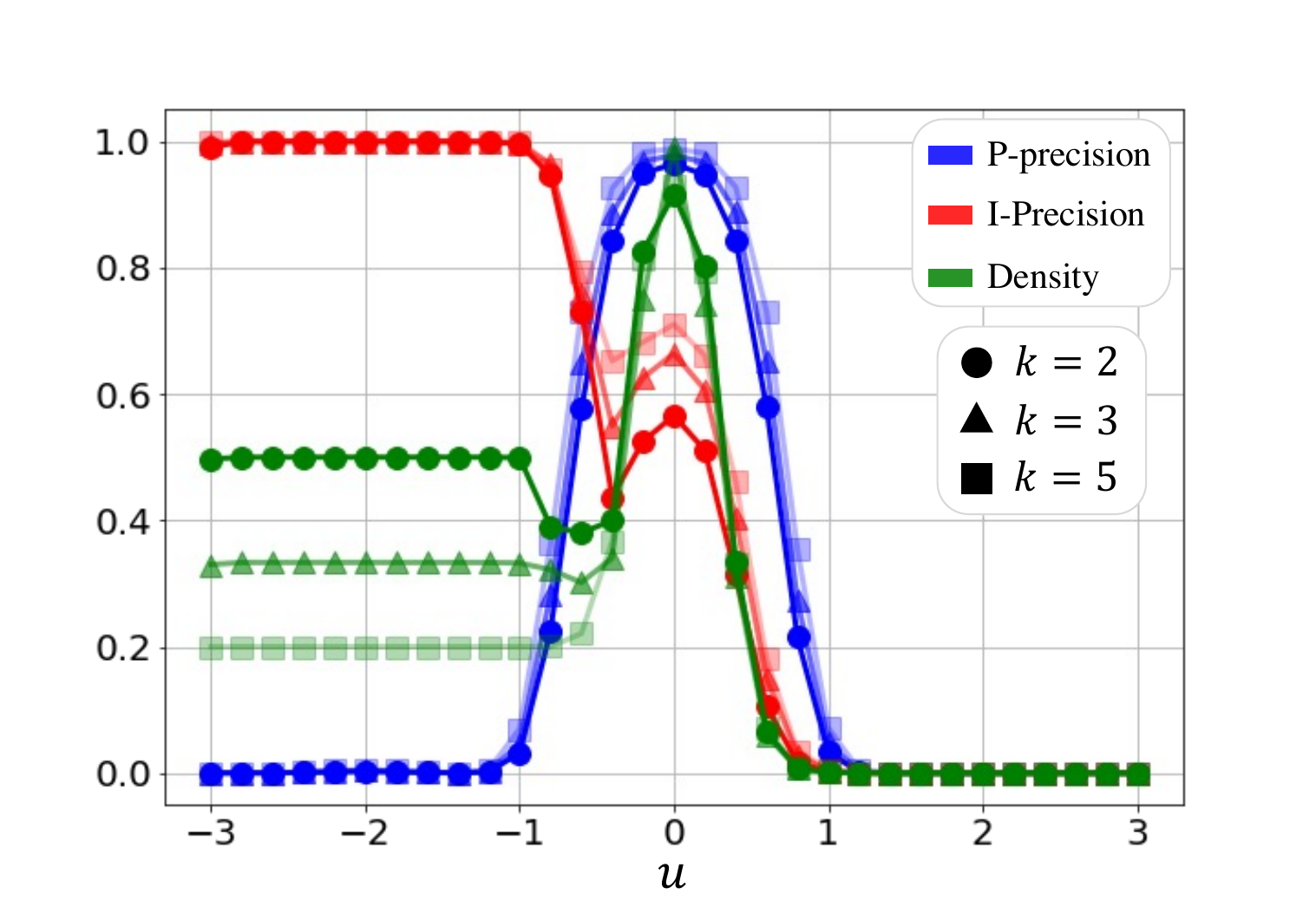}
\caption{Ablation over $k$. We measure each metric for different $k$ between $X\sim N(0,I)$ and $Y \sim N(u,I)$ as $u \in$ [-3.0,3.0] with outlier $x_o \sim N(-2,I)$ added to $X$.}
\label{fig:main3_1}
\end{figure}

It is also important to select $R$ in Eq.~\ref{eq:main10} since it controls the robustness of the scoring rule. Instead of sample-specific $R(x_i)$ based on $k$NN as used in IP\&IR and D\&C, we propose to use identical $R$ for all samples $x_i$. 
We set $R$ as a multiple of the average of $k$NN distances among $\{x_i\}_{i=1}^N$, 
since it can be used as an effective normalization factor in high-dimensional space. 
This is similar to using a fixed-variance Gaussian kernel for Kernel Density Estimation and offers several advantages over sample-specific $k$NN: 1) it suppresses the overestimation of $k$NN by the outlier as shown in Fig.~\ref{fig:main1}b, and 2) makes metric more robust to the choice of $k$ as shown in Fig.~\ref{fig:main3_1}.  

Finally, putting it all together, we propose P-precision as
\begin{equation}
\begin{split}
    \text{P-precision} = \frac{1}{M} \sum_{j=1}^M \text{PSR}_P(y_j)
\end{split}
\end{equation}
where
\begin{equation*}
\begin{split}
    \text{PSR}_P(y_j) = 1- \prod_{i=1}^N \left(1-\text{Pr}(y_j \in s_P(x_i))\right),
\end{split}
\end{equation*} 
\begin{equation*}
\begin{split}
    \text{Pr}(y_j \in s_P(x_i)) = 
    \begin{cases}
    1 - \frac{||x_i-y_j||_2}{R}, & \text{if} \ ||x_i-y_j||_2 \leq R \\
    0, & \text{otherwise}
    \end{cases} \\ 
\end{split}
\end{equation*}
and
\begin{equation*}
    R = \frac{a}{N}\sum_{i=1}^N \text{NND}_k(x_i).
\end{equation*}
$a$ is a hyperparameter that controls the magnitude of $R$, and $k$ is for the $k$-nearest neighbors. 
In a similar manner, we propose P-recall as
\begin{equation}
\begin{split}
    \text{P-recall} = \frac{1}{N} \sum_{i=1}^N \text{PSR}_Q(x_i)
\end{split}
\end{equation}
where
\begin{equation*}
\begin{split}
    \text{PSR}_Q(x_i) = 1- \prod_{j=1}^M \left(1-\text{Pr}(x_i \in s_Q(y_j))\right),
\end{split}
\end{equation*} 
\begin{equation*}
\begin{split}
    \text{Pr}(x_i \in s_Q(y_j)) = 
    \begin{cases}
    1 - \frac{||x_i-y_j||_2}{R}, & \text{if} \ ||x_i-y_j||_2 \leq R \\
    0, & \text{otherwise}
    \end{cases} \\ 
\end{split}
\end{equation*}
and
\begin{equation*}
    R = \frac{a}{M}\sum_{j=1}^M \text{NND}_k(y_j).
\end{equation*}

%% file: Sections/4_Experiments.tex
\begin{figure*}[t]
\centering
\subfloat[FFHQ]{
\includegraphics[width=0.66\textwidth]{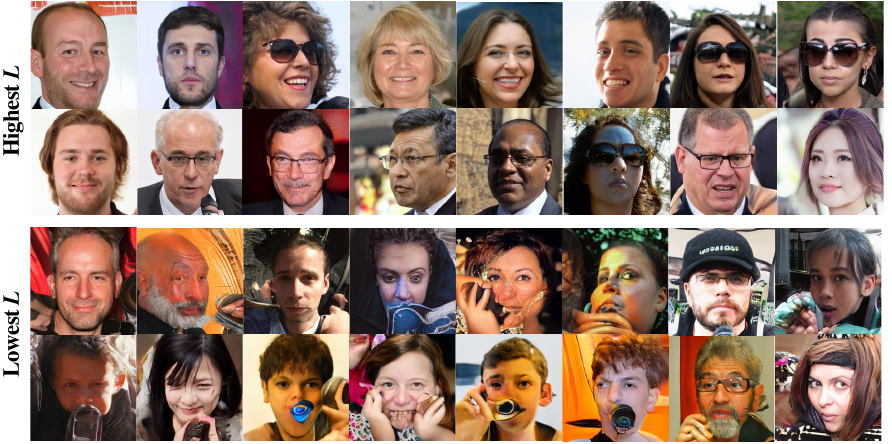}
\label{fig:main3_a}}
\subfloat[CIFAR-10]{
\includegraphics[width=0.27\textwidth]{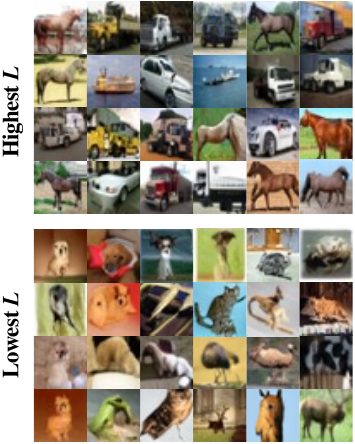}
\label{fig:main3_b}}
\caption{Qualitative examples sorted according to $L$. We used StyleGAN \cite{karras2019style} trained on FFHQ \cite{karras2019style} and BigGAN \cite{brock2018large} trained on CIFAR-10 \cite{krizhevsky2012imagenet}. 
The top two rows are images with the highest $L$ meaning they have high PSR but low DSR. Conversely, the bottom two rows are images with the lowest $L$, indicating low PSR and high DSR.
}
\label{fig:main3}
\end{figure*}

\section{Experiments}\label{sec:experiment}
In this section, we systematically demonstrate the limitations of current metrics and the improvements of our metrics through toy experiments. We then show the experiments with state-of-the-art generative models, highlighting that our metrics are more reliable for evaluating generative models than IP\&IR and D\&C. We use the settings described below for the remainder of the experiments unless stated otherwise. We set $k=4$ and $a=1.2$ for our metrics in all experiments. For IP\&IR and D\&C, we follow the recommended value of $k=3$ and $k=5$, respectively.
In the toy experiments, we generate Gaussian distributions in $\mathbb{R}^\text{64}$ and set $M=N=\text{10000}$. For the experiments evaluating generative models, we use features extracted from the ImageNet~\cite{krizhevsky2012imagenet} pre-trained Inception model \cite{szegedy2016rethinking} after the second fully connected layer. We compute metrics between all available real samples (up to 50$\mathrm{K}$) and 50$\mathrm{K}$ generated images following ~\cite{dhariwal2021diffusion}.

\begin{figure}[t]
\centering
\subfloat[]{\includegraphics[width=0.495\columnwidth]{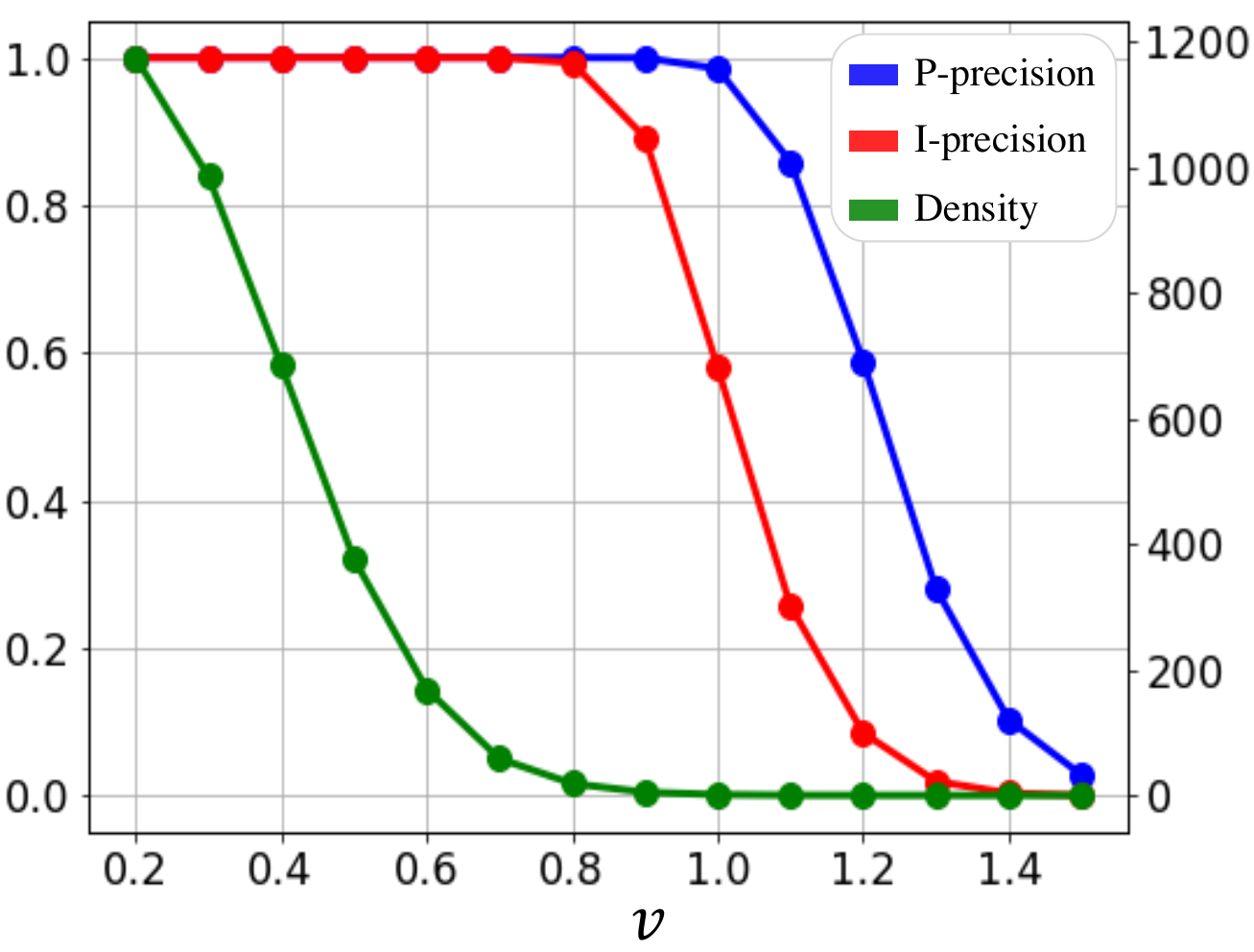}
\label{fig:main4_a}}
\subfloat[]{\includegraphics[width=0.495\columnwidth]{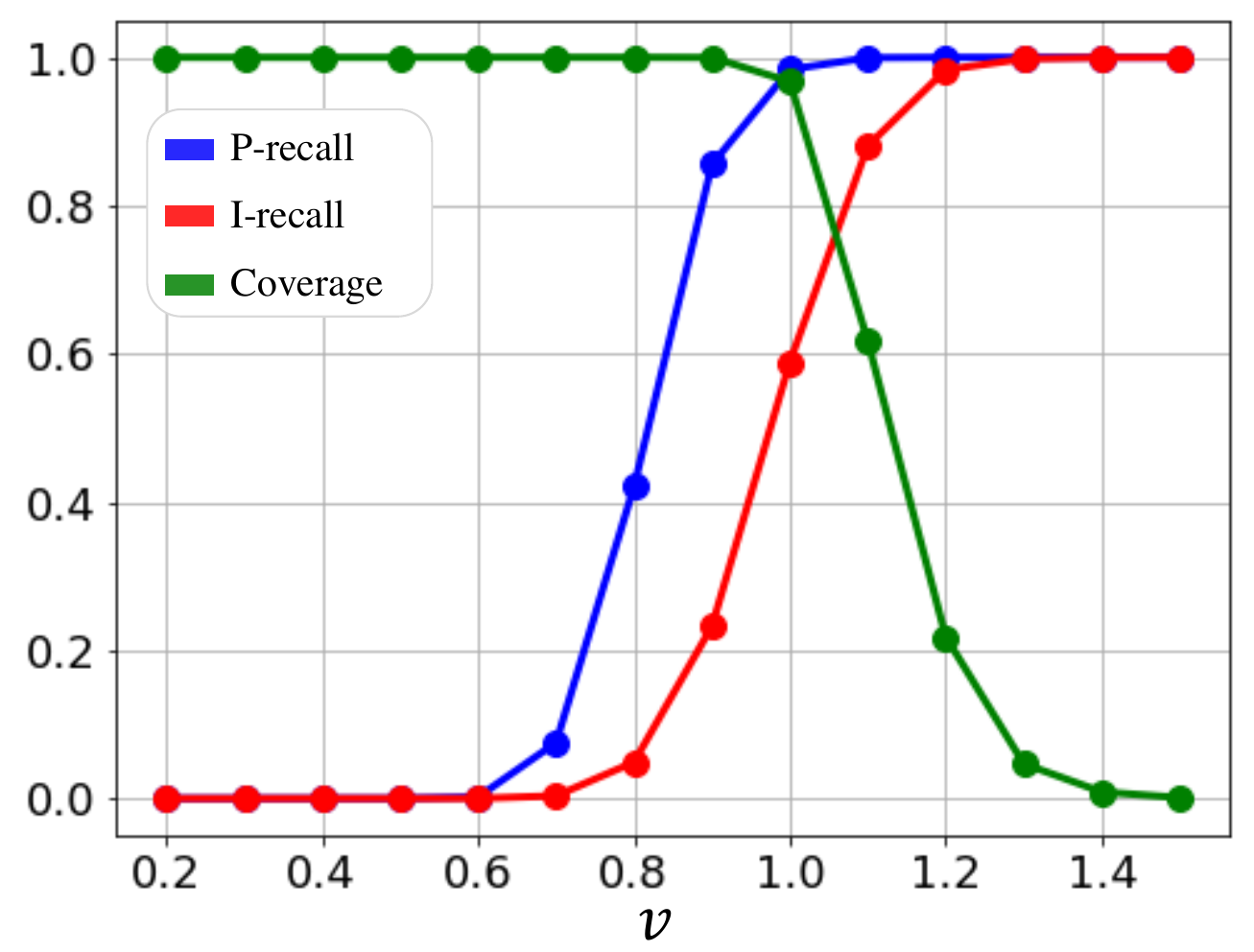}
\label{fig:main4_b}}
\caption{Behavior of metrics between $X \sim N(0,I)$ and $Y \sim N(0, vI)$ as $v$ changes between [0.2, 1.5]. Because Density goes over 1 (up to nearly 1000), the $y$-axis for Density is on the right side of the plot for better visualization.}
\label{fig:main4}
\end{figure}

\subsection{Toy experiments}
\subsubsection{Robustness to outliers}\label{sec:experiment_outlier}
\paragraph{Gaussian distribution}
We first investigate how the metrics behave to outliers through a simple toy experiment with Gaussian distributions, following \cite{naeem2020reliable}. We set real $X \sim N(0,I)$ and fake $Y \sim N(u\textbf{1},I)$ where \textbf{1} is a vector of ones and $I$ is the identity matrix, and measure the metrics between $X$ and $Y$ as $u$ varies in [-3, 3]. To introduce an outlier, we add a sample $x_o \sim N(\textbf{-2}, I)$ to the set of real samples. A reliable metric should consistently decrease as $u$ moves away from 0, regardless of the outlier. Fig.~\ref{fig:main2}a shows the result for fidelity: I-precision gives overestimated values when $u \leq 0$ due to the enlarged constant-density hypersphere by the outlier. Although Density alleviates the problem by reducing the density inside the hypersphere by $1/k$, it does not reflect the change for a wide range of $u$ (See when $u \leq$ -1). On the other hand, P-precision denotes reliable estimates even when the fake distribution comes closer to the outlier (PP = 0.006 when $u$ = -2), and also shows a robust decrease as $u$ moves away from 0. We observe a similar result on the diversity measures except for Coverage (See Fig.~\ref{fig:main2}b). 

In addition, we perform an ablation over $k$ to investigate whether the improvement of our metric over the outlier is due to a properly chosen $k$. The same experiment described above is repeated with different $k$ and the metric values are reported in Fig.~\ref{fig:main3_1}. Our P-precision shows strong consistency over $k$, due to the employment of identical $R$ for all real samples. However, I-precision continues to suffer from outliers, regardless of the value of $k$. Moreover, Density exhibits high variation over different values of $k$ as it reduces the density by $1/k$, demonstrating that Density needs properly chosen $k$ to achieve adequate robustness to outliers. 

These results suggest that our metrics are more reliable to estimate fidelity and diversity between distributions, even in the presence of outliers.

\subsubsection{Stability of metrics}
We also test the stability of metrics by measuring the empirical bias of the metric from the presumed true value, which we approximate by averaging across 50 runs with 50K samples, between two identical Gaussian distributions. Fig.~\ref{fig:main2}c shows the bias from the true value with respect to the number of datasets used for estimation, along with the average and standard deviation across 50 runs. Our PP\&PR show a small bias and standard deviation, even for small $N$, while IP\&IR show a large bias. In addition, despite the theoretical value when two distributions are identical \cite{naeem2020reliable}, Density does not behave nicely: at $N \leq 10000$, Density shows quite a large bias and standard deviation due to the extremely wide bound. This demonstrates that our PP\&PR provides more reliable estimates than other metrics even for small datasets.

\subsubsection{Reflecting fidelity and diversity}  \label{experiment:fd}
The main motive of the two-value metric is to capture both fidelity and diversity between two distributions. To assess whether these metrics are capable of reflecting fidelity and diversity, we design a scenario where the two values are at odds. Specifically, we generate samples from $X \sim N(0,I)$ and $Y \sim N(0,vI)$, and vary the value of $v$ from 0.2 to 1.5. As $v$ increases, we expect fidelity to decrease and diversity to increase. As shown in Fig.~\ref{fig:main4}, while Density behaves as expected, Coverage does not; it remains constant at 1 as $v$ increases from 0.2 to 1, despite the increasing relative diversity of fake distribution. We argue that this counter-intuitive behavior is due to the conceptual limitation of Coverage as discussed in Sec.~\ref{sec:background_problems}. When $v$ increases beyond 1, Coverage becomes 0, as fake samples are no longer included in the hyperspheres of real samples. In contrast, PP\&PR and IP\&IR nicely reflect the expected behavior. 

\subsection{Evaluating generative models}

\subsubsection{Comparing scoring rules}\label{sec:experiment_L}
Here, we compare the scoring rules of fidelity metrics to demonstrate how nicely the estimated density reflects the quality of the fake sample. We compare PSR with DSR, as BSR only denotes a binary score. For effective comparison, we focus on cases where two scoring rules have conflicting results. Specifically, we compute $L = \{\text{PSR}_P(y_j) - \text{DSR}_P(y_j) \mid \forall_j \}$ using generative models $Q(Y)$ (StyleGAN~\cite{karras2019style} and BigGAN~\cite{brock2018large}) trained on real-world datasets $P(X)$ (CIFAR-10~\cite{krizhevsky2012imagenet} and FFHQ~\cite{karras2019style}). A high value of $L$ indicates a high PSR but low DSR, while a low value of $L$ indicates the opposite. 
For a fair comparison, we normalize DSR by its maximum value so that it ranges between 0 and 1. We show images of the highest $L$ and lowest $L$ in Fig.~\ref{fig:main3} (without cherry-picking). In general, images with a high value of $L$ show more realistic and clear objects, while images with a low $L$ are often distorted with artifacts (See supplement for more examples). Based on the assumption that realistic images should correspond to the high density of real distribution, our results in Fig.~\ref{fig:main3} qualitatively demonstrate the superiority of PSR over DSR in reflecting the semantic quality of generated images. We also conduct a user study selecting preferences between the highest $L$ and the lowest $L$ and it also shows a similar result to our observation (See supplement).

\begin{figure}[t]
\centering
 \includegraphics[width=0.95\columnwidth]{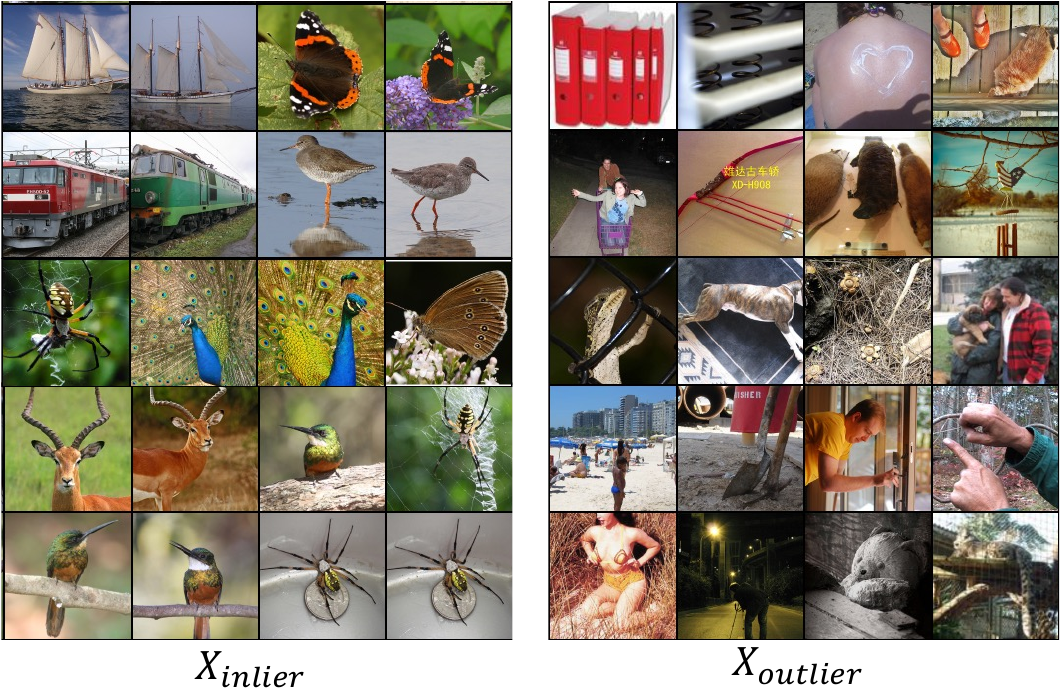} \\
\includegraphics[width=0.95\columnwidth]{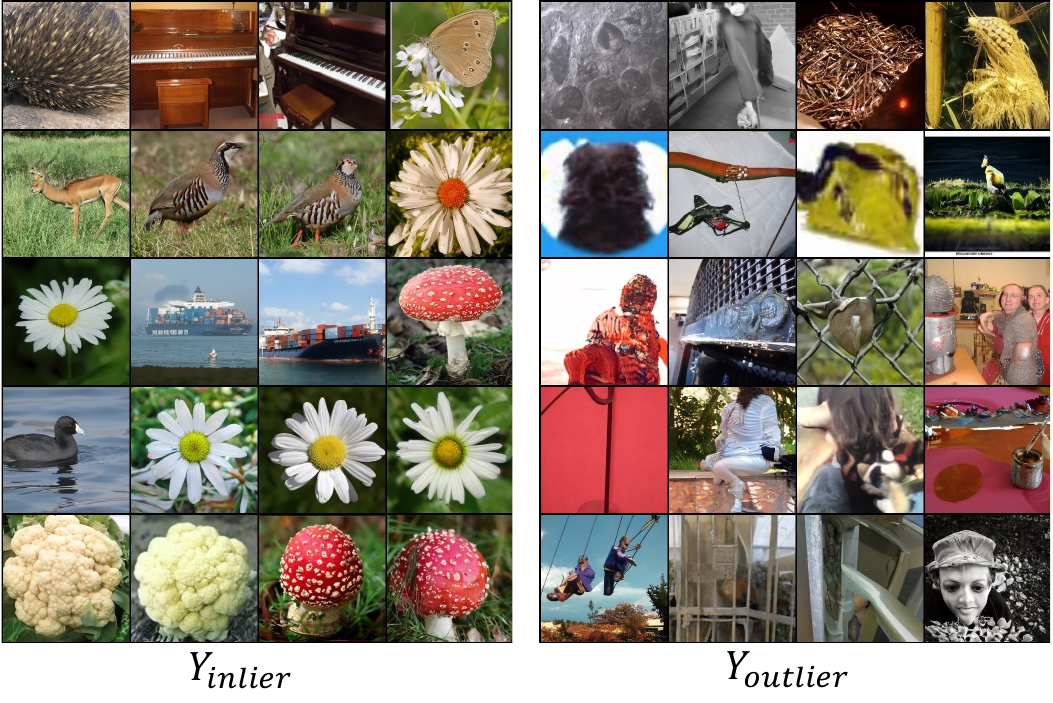} \\ 
\includegraphics[width=0.485\columnwidth]{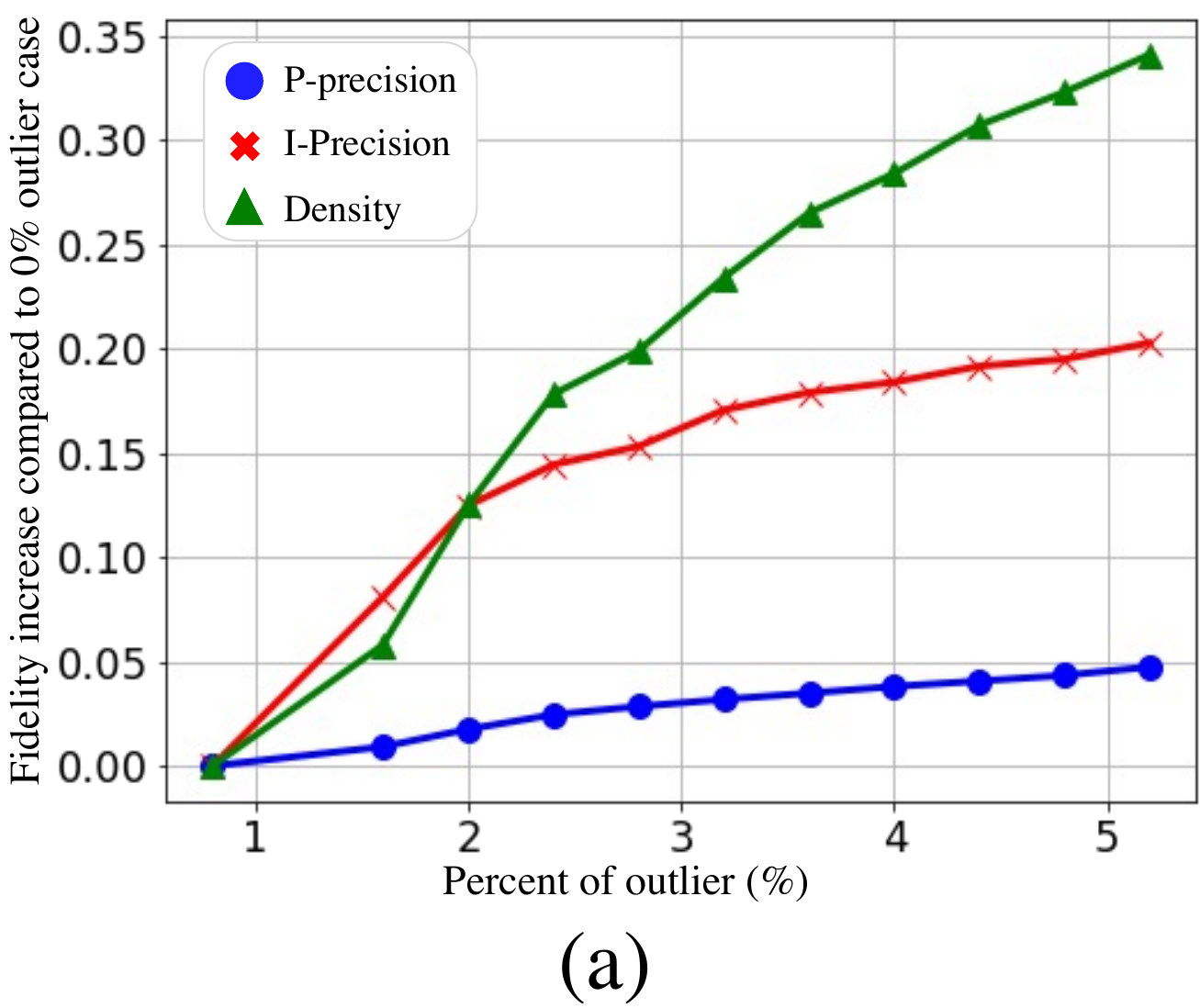}
\includegraphics[width=0.485\columnwidth]{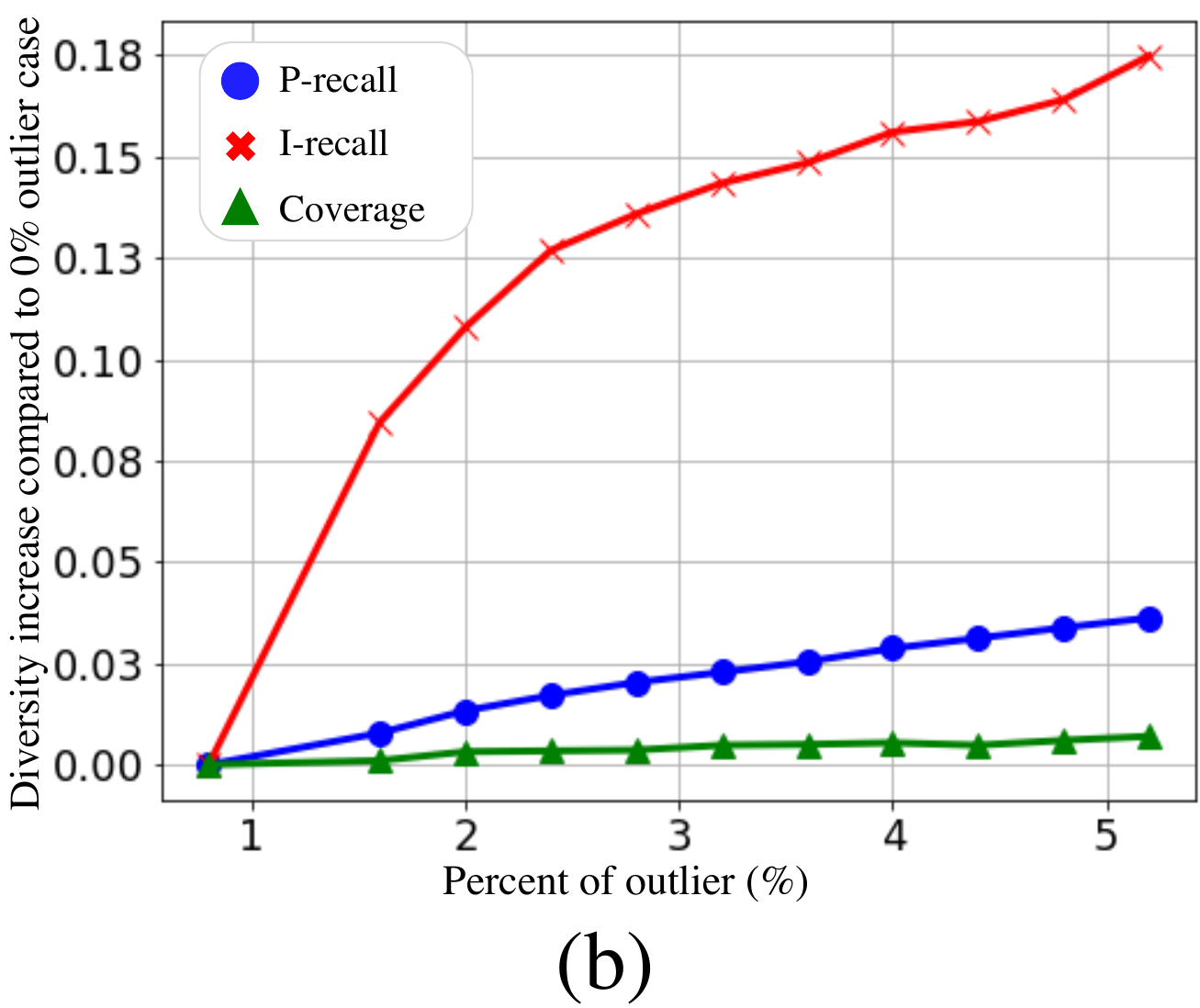}
\caption{Behavior of metrics with respect to the number of outliers when evaluating the generative model $Y$ on a real-world dataset $X$. (a) Fidelity metrics. (b) Diversity metrics.}
\label{fig:main5}
\end{figure}

\input{Table/table}

\subsubsection{Robustness to outliers}\label{sec:experiment_outlier_real}
To investigate the efficacy of our proposed metrics in handling outliers with generative models on real-world datasets, we conduct experiments on the state-of-the-art diffusion model (e.g., ADM~\cite{dhariwal2021diffusion}), which currently has gained attention for its powerful performance, trained on ImageNet~\cite{krizhevsky2012imagenet} dataset ($128^2$ resolution). Then, we adopt the outlier selection criterion used in ~\cite{naeem2020reliable} which
uses the distance to the $k^{th}$ nearest neighbor. We apply this criterion to select inliers and outliers (at a ratio of 0.95:0.05) for both real and fake samples generated by ADM, resulting in the sets $X=X_{inlier}\cup X_{outlier}$ and $Y=Y_{inlier}\cup Y_{outlier}$. We provide examples of the outliers and inliers in Fig.~\ref{fig:main5}. Note that the inliers tend to exhibit more recognizable objects with clear semantic features compared to the outliers.

We first observe how fidelity metrics between $X_{inlier}$ and $Y$ respond when a portion of $X_{inlier}$ is gradually substituted with the subset of $X_{outlier}$. To visualize this, we plot the metric increments compared to the case when no outliers are included in $X_{inlier}$, as shown in Fig.~\ref{fig:main5}a. The results show that I-precision rapidly increases (0.2 increments) when only 5\% of inliers are gradually replaced by outliers, confirming its vulnerability to outliers. Although Density is less affected by outliers initially, it eventually surpasses the increment of I-precision as it accumulates the density over hyperspheres, supporting our claim that Density is still vulnerable to outliers. On the other hand, our P-precision shows the least increment (0.05 increments for 5\% outliers), validating its effectiveness in handling outliers.

Likewise, we also test the robustness of diversity measures between $X$ and $Y_{inlier}$ by gradually replacing a portion of $Y_{inlier}$ with the subset of $Y_{outlier}$. In Fig.~\ref{fig:main5}b, we observe that I-recall exhibits a similar vulnerability to outliers as I-precision, while our P-recall is less affected by the presence of outliers. Besides, Coverage demonstrates strong robustness to outliers due to its alternative diversity measure. However, this comes at the cost of losing accuracy, as demonstrated in the experiments in Fig.~\ref{fig:main4}b and ~\ref{fig:main6}b.

\subsubsection{Reflecting fidelity and diversity}
\begin{figure}[t]
\centering
\includegraphics[width=0.495\columnwidth]{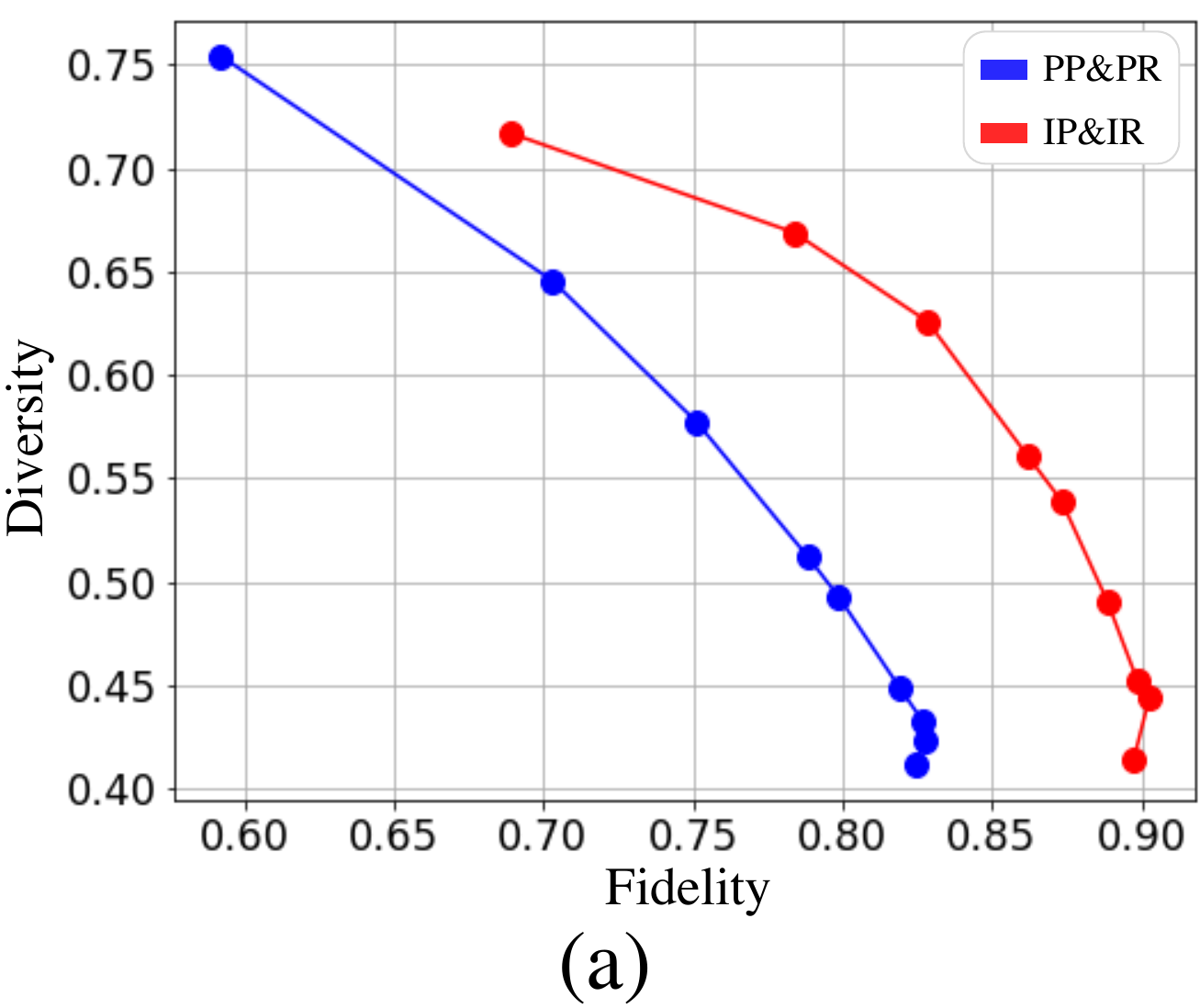}
\includegraphics[width=0.495\columnwidth]{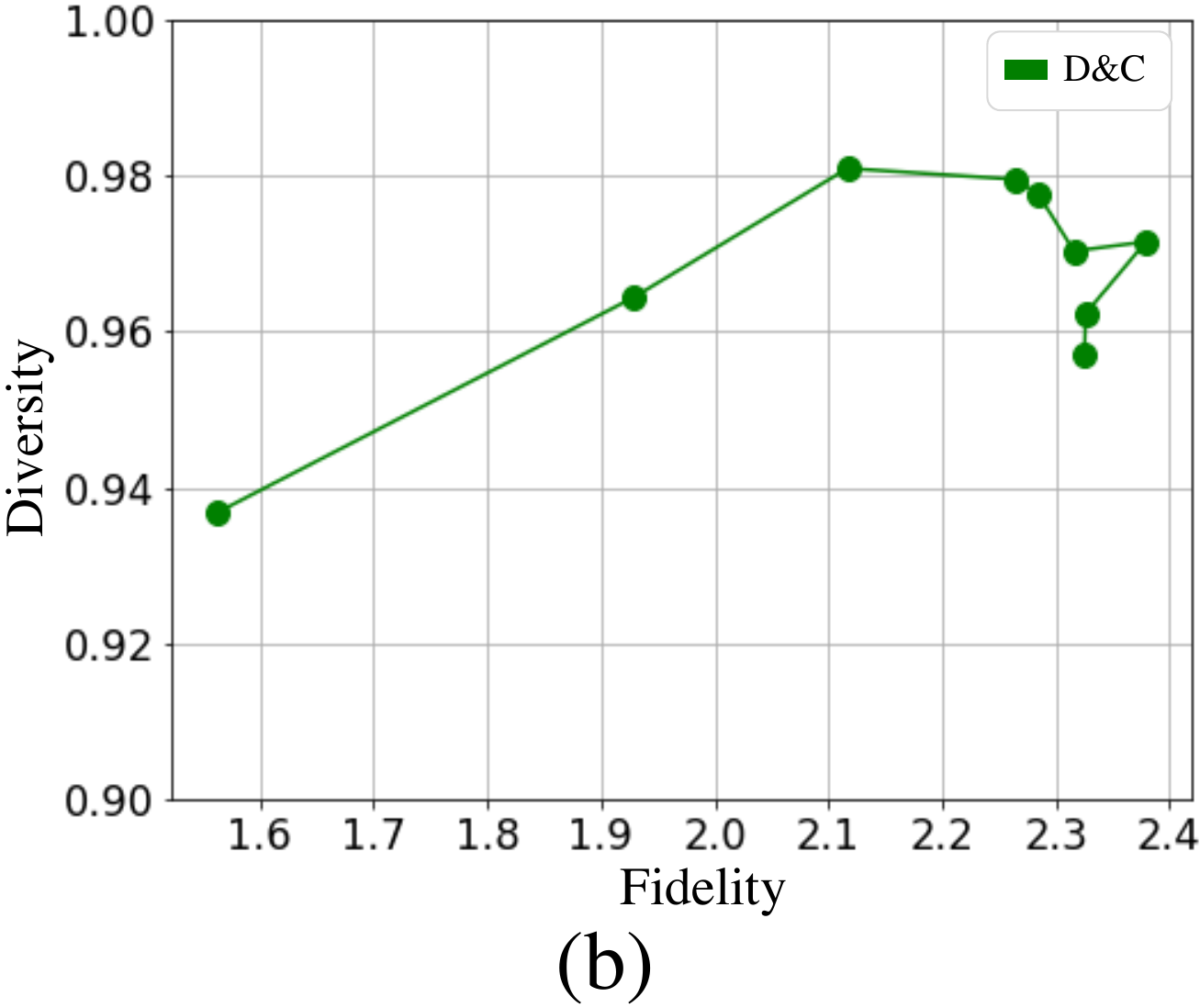}
\caption{Fidelity versus diversity when increasing the gradient scale from 0.5 to 10.0 for sampling diffusion models with classifier guidance \cite{dhariwal2021diffusion}. A larger gradient scale tends to result in higher fidelity and lower diversity.}
\label{fig:main6}
\end{figure}

We now verify whether the metrics accurately reflect the change in fidelity and diversity of fake samples generated from generative models. In order to do that, we employ classifier guidance ~\cite{dhariwal2021diffusion} during the image sampling from ADM. This technique uses gradients of a pre-trained classifier to perform conditional sampling from pre-trained unconditional diffusion models. By adjusting the scale of gradients from the classifier, we can explicitly control the balance between fidelity and diversity of generated images. For instance, a larger gradient scale results in the diffusion model sampling more from the modes of the pre-trained classifier, yielding higher fidelity but lower diversity. For more details, please refer to ~\cite{dhariwal2021diffusion}. Therefore, we sample images from ADM, the same model as in Sec.~\ref{sec:experiment_outlier_real}, using gradient scales ranging from 0.5 to 10.0, and measure metrics (IP\&IR, D\&C and PP\&PR) between real and generated samples for each scale. Fig.~\ref{fig:main6} shows the results; our PP\&PR and IP\&IR effectively capture the trade-off between fidelity and diversity whereas D\&C does not exhibit a clear relationship with the gradient scale. This is due to the conceptual limitation of Coverage in accurately measuring the diversity in generated samples.

\subsubsection{Image generation benchmark}
To evaluate the practical usefulness of our proposed metrics in assessing the state-of-the-art generative models, we present the quantitative results on real-world datasets (LSUN bedroom~\cite{yu2015lsun}, AFHQ-v2~\cite{choi2020stargan}, and ImageNet) for various types of generative models, along with FID. The results, presented in Tab.~\ref{tab:main2}, are obtained by measuring each metric five times and taking their average. For clarity in the presentation, the variance associated with these measurements can be found in the supplementary materials. While FID serves as a holistic reference point for distinguishing superior model performance, our metrics delve deeper, elaborating on aspects of fidelity and diversity. For instance, our metrics explain that the improvement of ADM over StyleGAN~\cite{karras2019style} on LSUN bedroom in terms of FID is due to an increase in fidelity at the cost of diversity. An analysis between ADM and ADM-G further reveals that the classifier guidance in ADM-G augments fidelity, yet curtails diversity. This demonstrates the usefulness of our metrics in elaborating on the effect of the specific component in models.

Moreover, to observe whether the metrics harmonize with the widely adopted FID metric, we analyze the consistency between FID and two-value metrics. This is achieved by computing the $\text{F}_1$-score~\cite{goutte2005probabilistic}, which is a harmonic mean of fidelity and diversity. The results demonstrate that $\text{F}_1$-scores of our metric mirror the trends set by FID (as underscored by the bold scores for each dataset in Tab.~\ref{tab:main2}).

\begin{figure}
\centering
\includegraphics[width=0.72\columnwidth]{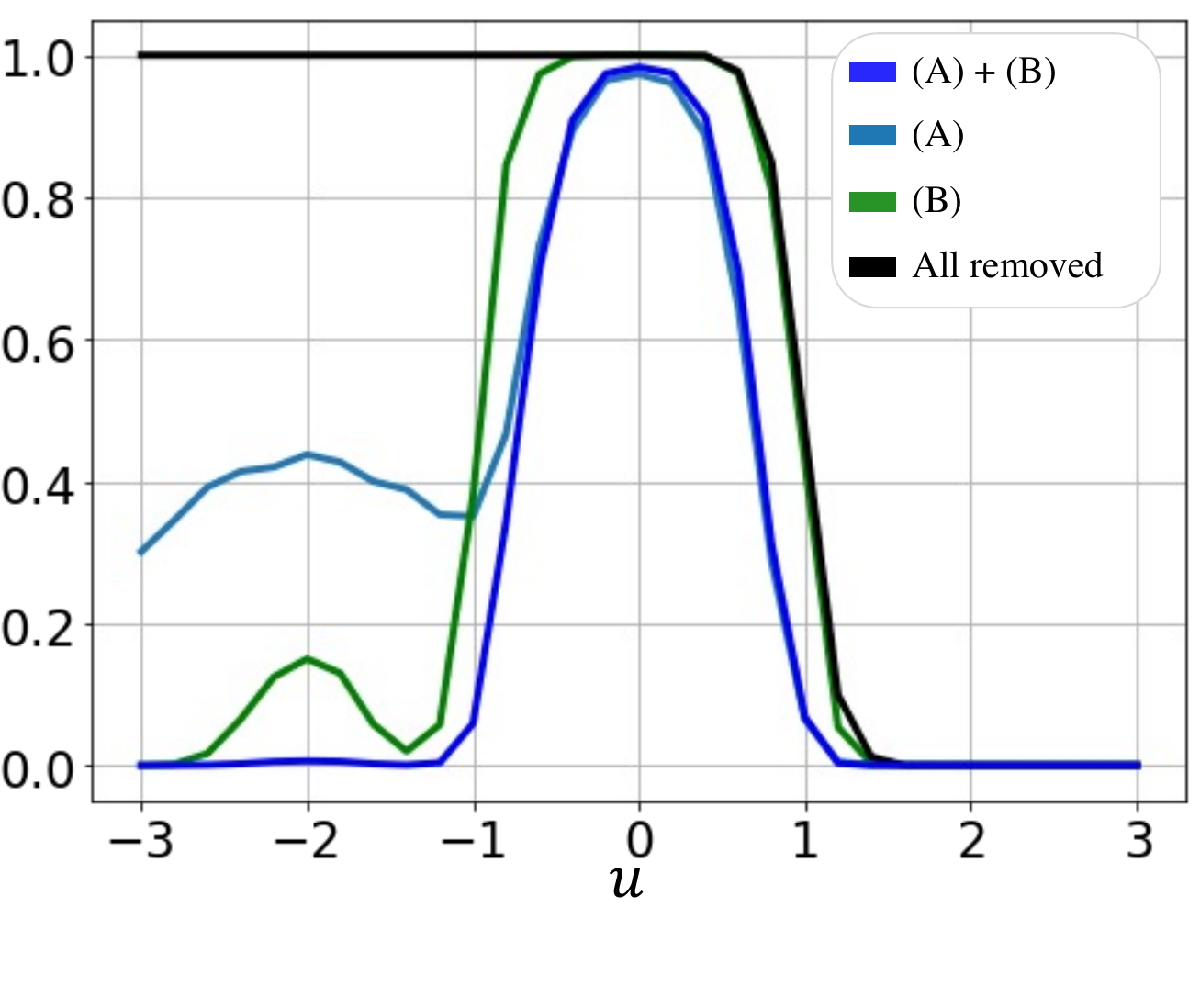}
\caption{Ablation study on PSR: (A): probabilistic estimation for subsupport. (B): identical $R$.}
\label{fig:main7}
\end{figure}

\section{Ablation study}\label{section:ablation}
Here, we conduct an ablation study on our proposed scoring rule, PSR, to demonstrate the effectiveness of our approach: (A) probabilistic estimation for subsupport (Eq.~\ref{eq:main10}) and (B) identical $R$. We repeat the outlier toy experiment with Gaussian distributions in Sec.~\ref{sec:experiment_outlier} with different configurations and the result for fidelity metric is shown in Fig.~\ref{fig:main7}. In both cases where the feature (A) and (B) are individually removed, the metric becomes more susceptible to outlier than the case where both are used ((A)+(B)) but less affected than the case where both are all removed (see when $u\in$[-3,-1]). This demonstrates that our probabilistic approach for subsupport and addressing overestimation of sample-specific $k$NN both contribute to estimating accurate density and combining them together maximizes it.

\section{Conclusions and limitations}
In this paper, we introduced a set of two-value metrics, P-precision and P-recall, based on the probabilistic approach that addresses the limitations of existing metrics (IP\&IR and D\&C). We have conducted a systematic investigation of the limitations and demonstrated the effectiveness of our proposed metrics by evaluating toy experiments and state-of-the-art generative models. Nevertheless, we recognize there is still room for improvement. While we utilized commonly-used ImageNet embeddings~\cite{heusel2017gans, kynkaanniemi2019improved,naeem2020reliable} for feature extraction, recent studies such as \cite{alaa2022faithful, kynkaanniemi2022}, have proposed domain-agnostic embeddings to enhance the feature quality compared to previous embeddings that rely on dataset-specific statistics. Thus, future research focusing on the quality of feature embedding combined with our method may further enhance the reliability and usefulness of the metric.

\paragraph{Acknowledgements} This work was supported by the Institute of Information \& Communications Technology Planning \& Evaluation (IITP) grant funded by the Korea government(MSIT) (No.2021-0-00456, Development of Ultra-high Speech Quality Technology for remote Multi-speaker Conference System and No.RS-2023-00225630, Development of Artificial Intelligence for Text-based 3D Movie Generation).

%% file: Table/table.tex
\begin{table*}[t]
\caption{Quantitative result of various generative models on real-world datasets. The reported values are obtained by measuring each metric five times and taking the average. Bolded values indicate the top scores evaluated by the corresponding metric.}
\centering
\resizebox{0.9\textwidth}{!}{%
\begin{tabular}{l|c|ccc|ccc|ccc} 
\hline
\multicolumn{1}{l}{Model} & \multicolumn{1}{c}{FID$\downarrow$} & PP$\uparrow$                   & PR$\uparrow$                   & \multicolumn{1}{c}{$\text{F}_1$$\uparrow$} & IP$\uparrow$                   & IR$\uparrow$                   & \multicolumn{1}{c}{$\text{F}_1$$\uparrow$} & D$\uparrow$                    & C$\uparrow$                    & $\text{F}_1$$\uparrow$              \\
\multicolumn{11}{l}{\textbf{ImageNet 256x256}} \\ 
\hline
\hline
ADM~\cite{dhariwal2021diffusion}                & 4.95                   & 0.538                & \textbf{0.732}               & 0.621                        & 0.681       & \textbf{0.688}                & 0.684                        & 1.52      & 0.876       & 1.112        \\
ADM-G~\cite{dhariwal2021diffusion}               & \textbf{4.58}          & 0.699                & 0.587       & \textbf{0.638}               & 0.818                & 0.606       & \textbf{0.696}               & 2.071                & \textbf{0.956}              & 1.307                 \\
BigGAN~\cite{brock2018large}               & 8.12                   & \textbf{0.751}       & 0.465                & 0.574                        & \textbf{0.874}                & 0.403                & 0.551                        & \textbf{2.481}                & 0.945                & \textbf{1.368}                \\
\multicolumn{1}{l}{}      & \multicolumn{1}{l}{}    & \multicolumn{1}{l}{} & \multicolumn{1}{l}{} & \multicolumn{1}{l}{}         & \multicolumn{1}{l}{} & \multicolumn{1}{l}{} & \multicolumn{1}{l}{}         & \multicolumn{1}{l}{} & \multicolumn{1}{l}{} & \multicolumn{1}{l}{}  \\
\multicolumn{11}{l}{\textbf{AFHQv2 512x512}} \\ 
\hline
\hline
StyleGAN2~\cite{karras2020analyzing}                 & 4.62                    & \textbf{0.568}       & 0.689                & 0.623                        & \textbf{0.716}       & 0.494                & 0.584                        & \textbf{1.886}       & 0.790                & \textbf{1.113}        \\
StyleGAN3-R~\cite{karras2021alias}               & 4.40                    & 0.564                & 0.725                & 0.634                        & 0.685                & \textbf{0.591}       & \textbf{0.635}               & 1.576                & 0.770                & 1.034                 \\
StyleGAN3-T~\cite{karras2021alias}               & \textbf{4.04}           & 0.567                & \textbf{0.727}       & \textbf{0.637}               & 0.699                & 0.578                & 0.632                        & 1.624                & \textbf{0.792}       & 1.065 \\
\multicolumn{1}{l}{}      & \multicolumn{1}{l}{}    & \multicolumn{1}{l}{} & \multicolumn{1}{l}{} & \multicolumn{1}{l}{}         & \multicolumn{1}{l}{} & \multicolumn{1}{l}{} & \multicolumn{1}{l}{}         & \multicolumn{1}{l}{} & \multicolumn{1}{l}{} & \multicolumn{1}{l}{}  \\
\multicolumn{11}{l}{\textbf{LSUN Bedroom 256x256}} \\ 
\hline
\hline
DDPM~\cite{karras2020analyzing}                 & 4.88                    & 0.799       & 0.749                & 0.773                        & 0.606       & 0.444                & 0.512                        & 1.701       & 0.977                & 1.241        \\
ADM~\cite{dhariwal2021diffusion}               & \textbf{1.91}                    & \textbf{0.839}                & 0.731                & \textbf{0.781}                        & \textbf{0.659}                & 0.494       & \textbf{0.565}               &\textbf{1.929}                & \textbf{0.993}                & \textbf{1.311}                 \\
StyleGAN2~\cite{karras2020analyzing}               & 2.35           & 0.801                & \textbf{0.753}       & 0.776               & 0.591                & \textbf{0.501}                & 0.543                        & 1.732                & 0.986       & 1.257 \\
\multicolumn{1}{l}{}      & \multicolumn{1}{l}{}    & \multicolumn{1}{l}{} & \multicolumn{1}{l}{} & \multicolumn{1}{l}{}         & \multicolumn{1}{l}{} & \multicolumn{1}{l}{} & \multicolumn{1}{l}{}         & \multicolumn{1}{l}{} & \multicolumn{1}{l}{} & \multicolumn{1}{l}{}          
\end{tabular}
}\label{tab:main2}
\end{table*}

%% file: Sections/5_Appendix.tex
\section{Proof}
\subsection{Proof of PSR bounded between 0 and 1}
\begin{proof}
In the main paper, our proposed scoring rule $\text{PSR}_P$ is defined as
\begin{equation*}
\begin{split}
    \text{PSR}_P(y_j) = 1- \prod_{i=1}^N \left(1-\text{Pr}(y_j \in S_P(x_i))\right),
\end{split}
\end{equation*} 
where
\begin{equation*}\label{equation:4}
\begin{split}
    \text{Pr}(y_j \in S_P(x_i)) = 
    \begin{cases}
    1 - \frac{||x_i-y_j||_2}{R}, & \text{if} \ ||x_i-y_j||_2 \leq R \\
    0, & \text{otherwise}
    \end{cases} \\ 
\end{split}
\end{equation*}
and
\begin{equation*}
    R = \frac{a}{N}\sum_{i=1}^N \text{NND}_k(x_i).
\end{equation*}
By the definition of $\text{Pr}(y_j \in S_P(x_i))$, it is bounded between 0 and 1 for all $i$:
\begin{align*}
        0 \leq \text{Pr}(y_j \in S_P(x_i)) \leq 1, \forall_i
\end{align*}
Thus,
\begin{align*}
        0 \leq 1-\text{Pr}(y_j \in S_P(x_i)) \leq 1, \forall_i
\end{align*}

Multiplication of $N$ values within [0,1] is also between 0 and 1 by the inequality properties:
\begin{align*}
    0 \leq \prod_{i=1}^N \left(1-\text{Pr}(y_j \in S_P(x_i)) \right) \leq 1
\end{align*}
Finally, 
\begin{align*}
    0 \leq \text{PSR}_P(y_j) = 1- \prod_{i=1}^N \left(1-\text{Pr}(y_j \in S_P(x_i)) \right) \leq 1
\end{align*}
Therefore, $\text{PSR}_P(y_j)$ is bounded within [0,1]. The same procedure symmetrically applies to $\text{PSR}_Q(x_i)$.

\end{proof}

\section{Related work}
\paragraph{Statistical divergence metrics} mainly measure the disparity between real and generated image distributions into a single value. For instance, the Inception Score~\cite{salimans2016improved} (IS) assesses the quality of generated images, utilizing the label distribution assigned by the InceptionV3 model. High diversity and label confidence in generated images result in a better score. The Frechet Inception Distance~\cite{heusel2017gans} (FID) computes the distance between real and generated image distributions in a feature space defined by the Inception network. A lower FID implies that generated images are closer to real ones in this feature space. Meanwhile, the Kernel Inception Distance~\cite{binkowski2018demystifying} (KID) refines the concept behind FID by relaxing its Gaussian assumption. Instead, KID calculates the squared Maximum Mean Discrepancy between Inception representations of real and generated samples with a polynomial kernel. Although these metrics are widely adopted due to their alignment with human evaluations, they fall short in providing detailed insights into fidelity and diversity.

\paragraph{Precision and recall metrics}
Precision and Recall \cite{sajjadi2018assessing, kynkaanniemi2019improved} introduced an approach using precision and recall to distinguish between fidelity and diversity of generative models. In this context, precision evaluates how closely generated samples resemble real ones, while recall gauges the generator's capability to reproduce all samples from the training set. Building on this foundation, subsequent research endeavored to address perceived limitations in the precision and recall framework. Density and Coverage \cite{naeem2020reliable} expressed concerns over the contemporary precision and recall metrics, highlighting that they:
1) struggle to recognize a perfect match between identical distributions,
2) are not sufficiently robust in the presence of outliers, and
3) rely on arbitrary evaluation hyperparameters. As a remedy, they introduced Density and Coverage metrics. While precision traditionally examines if a generated sample falls within any neighborhood sphere, Density evaluates the number of real-sample neighborhood spheres encompassing that generated sample. However, in this paper, we have systemically demonstrated that D\&C still remain vulnerable to outliers and have conceptual limitation that make Coverage insensitive to distribution change.

In a subsequent contribution, another work~\cite{alaa2022faithful} proposed a tri-dimensional evaluation measure, encompassing $\alpha$-Precision, $\beta$-Recall, and Authenticity, aimed at delineating the fidelity, diversity, and generalization prowess of generative models. They posit that a fraction $1-\alpha$ (or $1-\beta$) of real (or synthetic) samples can be considered as outliers, with the remainder deemed typical. Here, $\alpha$-Precision represents the proportion of synthetic samples that align with the most typical $\alpha$ real samples. Conversely, $\beta$-Recall indicates the proportion of real samples enveloped by the most representative $\beta$ synthetic samples. Importantly, $\alpha$-Precision and $\beta$-Recall span the entire [0, 1] interval, offering comprehensive precision-recall curves over a singular value. In order to employ their metric, they embed samples into hyperspheres, with a majority of samples clustered around the centers. Although their approach commendably addresses the outlier conundrum by adjusting metrics for varying supports, it relies on specific support modifications. This calls for extra training to create a specific embedding domain, such as a center-heavy hypersphere, culminating in computational challenges and practical constraints, particularly with expansive datasets like ImageNet. Contrarily, our proposed technique is a versatile solution, apt for any feature domain.

\section{Comparing scoring rules}
\subsection{User study}
We conducted a user study to evaluate the effectiveness of scoring rules in quantifying the quality of generated images. The study followed a 2-alternative force-choice paradigm as presented in previous work~\cite{saharia2022image}. Participants were presented with two images, one from the set $L_{low}$ which consisted of images with the lowest $L$ values as described in the main paper, and the other from the set $L_{high}$ which contained images with the highest $L$ values. Participants were then asked to select the image they considered to have better quality, i.e., more realistic and with fewer artifacts, based on the presented options.
The result presented in Tab.~\ref{tab:append1} affirms the effectiveness of our proposed scoring rule, PSR, in reflecting the quality of generated images.

\subsection{Additional examples}
We provide additional qualitative examples of generated images sorted by condition $L$ (PSR - DSS). Fig.~\ref{fig:append11} shows the generated images from BigGAN \cite{brock2018large} trained on CIFAR-10~\cite{krizhevsky2012imagenet}, sorted by condition $L$ without cherry-picking. Images with the highest $L$ tend to be more realistic than those with the lowest $L$. Fig.~\ref{fig:append12} also shows the generated images from StyleGAN \cite{karras2019style} trained on FFHQ \cite{karras2019style}, sorted by condition $L$ also without cherry-picking. Fig.~\ref{fig:append12}a shows images with the highest $L$ whereas Fig.~\ref{fig:append12}b shows images with the lowest $L$. Most images with high $L$ values are easily recognizable, with simple backgrounds. In contrast, most images with low $L$ values exhibit severe distortions, and their backgrounds are complicated and distorted. 

\input{Table/table_userstudy.tex}

\begin{figure}[t]
\centering
\subfloat[]{\includegraphics[width=0.41\columnwidth]{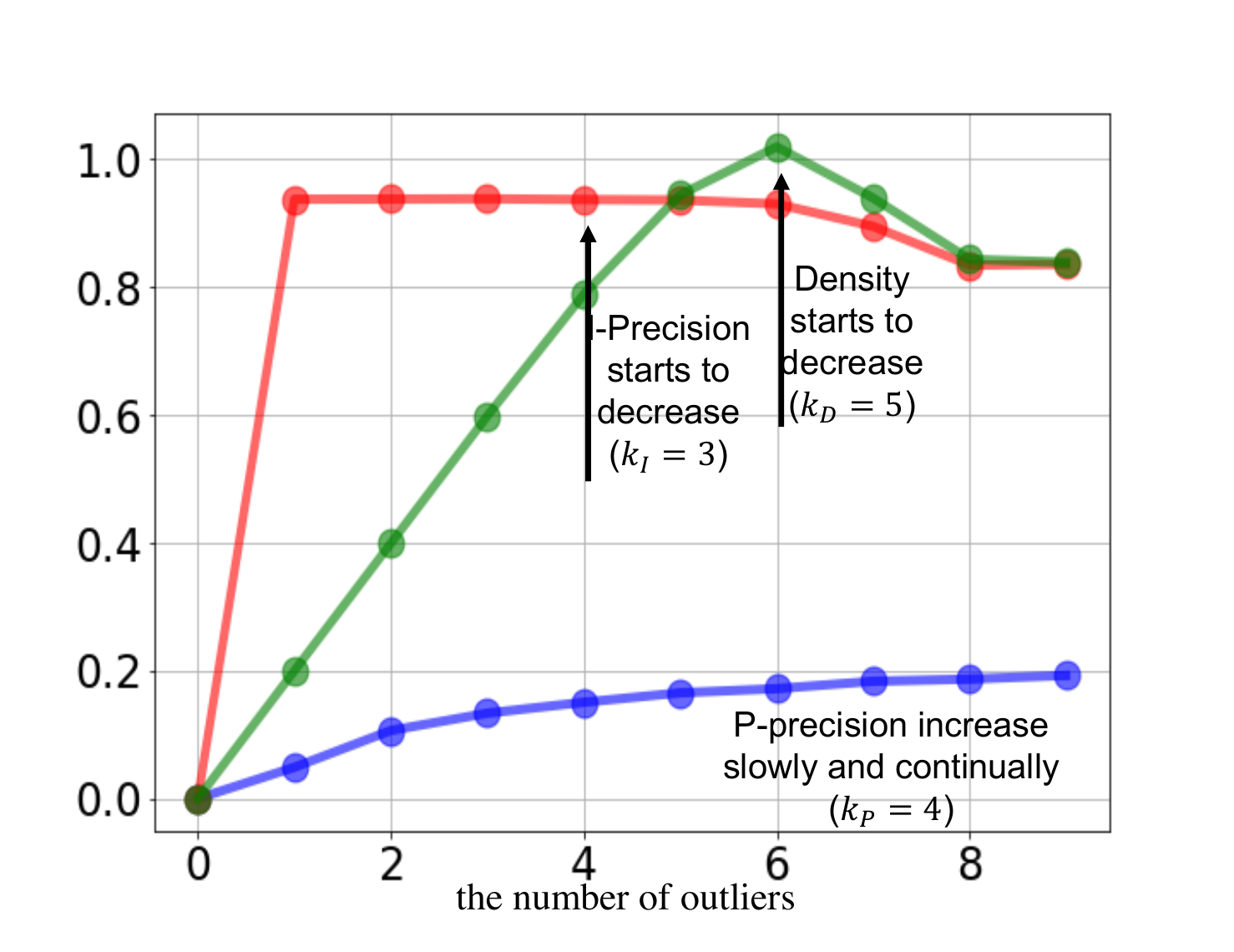}
}
\subfloat[]{\includegraphics[width=0.43\columnwidth]{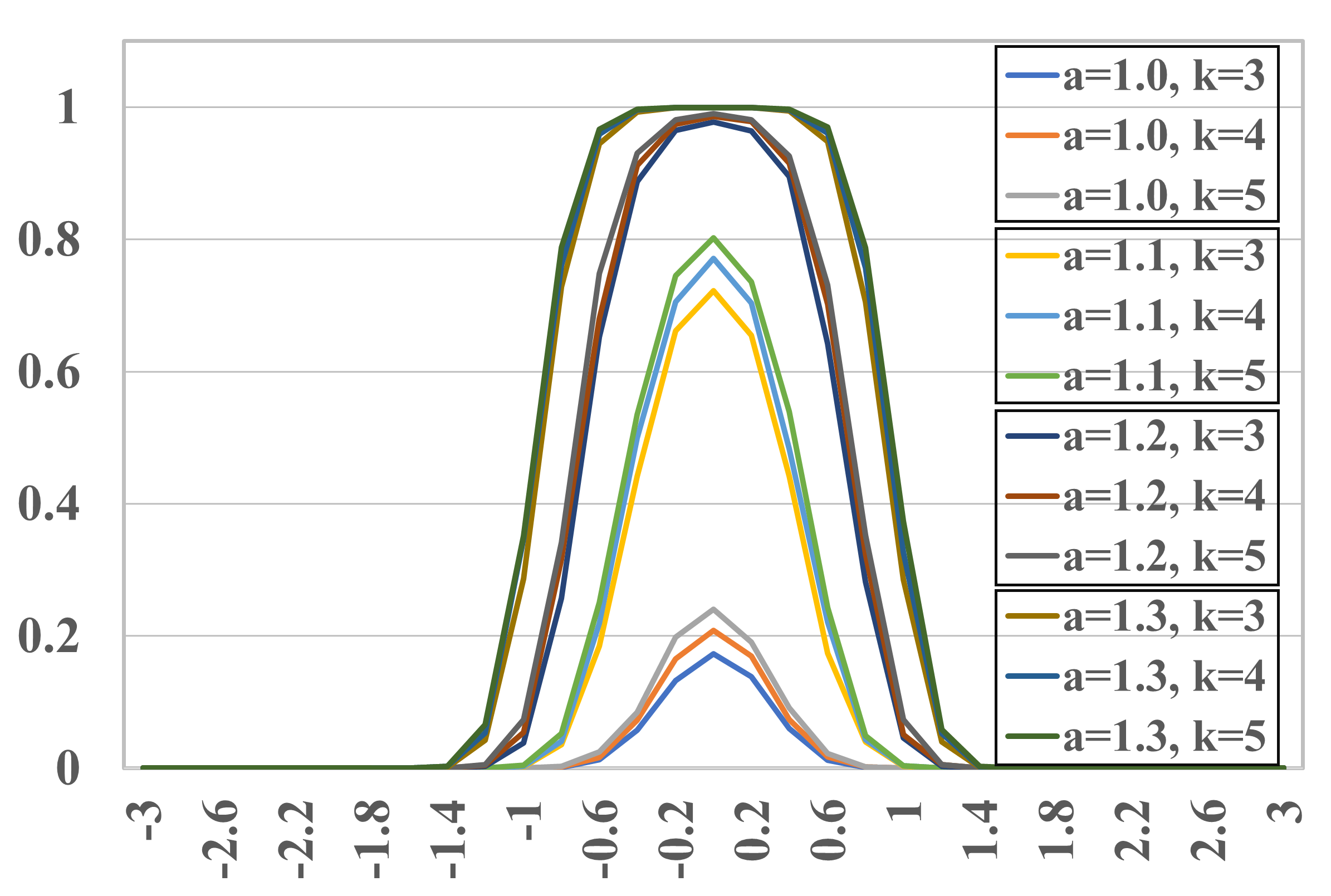}
}
\caption{(a) Behavior of metrics between $X \sim N(0,I)$ and $Y \sim N(0, vI)$ as $v$ changes between [0.2, 1.5]. Because Density goes over 1 (up to nearly 1000), the $y$-axis for Density is on the right side of the plot for better visualization. (b) Ablation study for selecting hyperparameter $a$ and $k$.}
\label{fig:append9}
\end{figure}

\begin{figure}[h]
\centering
\includegraphics[width=0.93\textwidth]{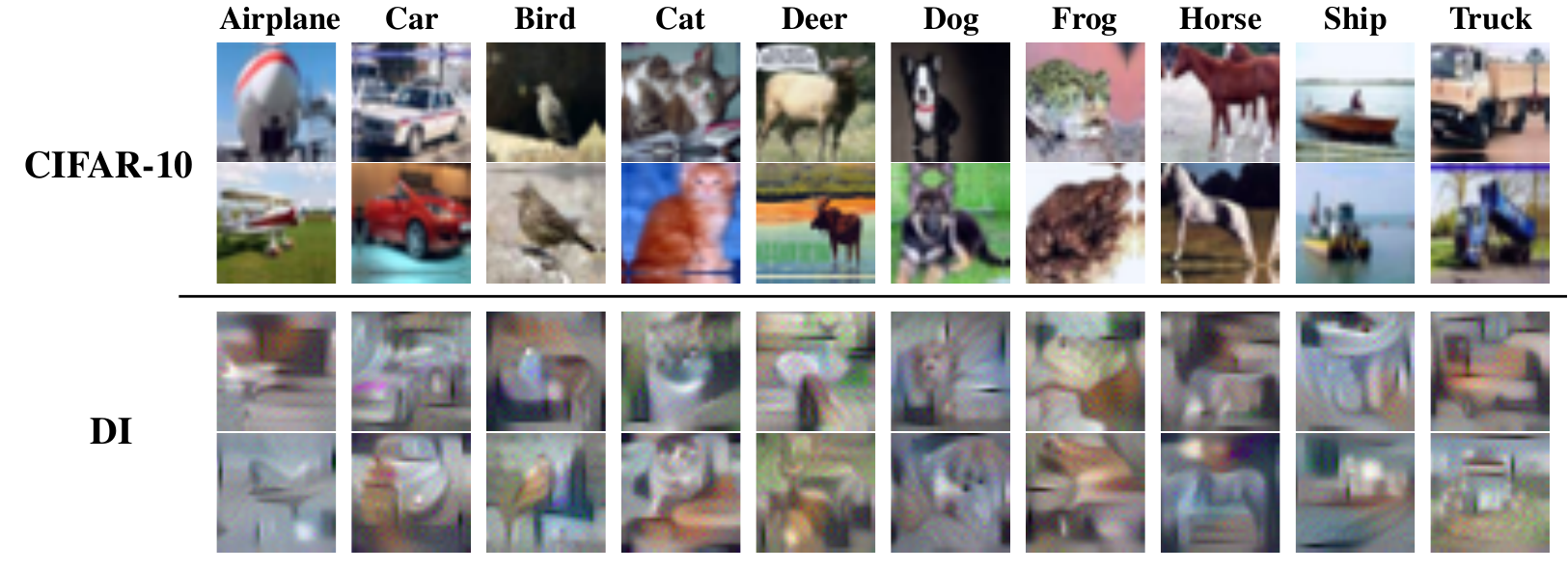}
\caption{Qualitative examples of CIFAR-10~\cite{krizhevsky2012imagenet} and the images generated by DI~\cite{yin2020dreaming}.}
\label{fig:append10}
\end{figure}

\section{Additional experiments}
\subsection{Hyperparameter selection}
Our metric has two defining hyperparameters:  $a$ and $k$.  In order to pinpoint hyperparameters that optimize both the robustness and sensitivity of the metric, we revisited the outlier experiment from Sec.~5.1 (in the main paper) for varying $a$ and $k$, and the results are shown in Fig.~\ref{fig:append9}b. 
An increase in $a$ diminishes the metric's sensitivity to shifts in distribution, a consequence of our scoring rule becoming more lenient (the slope of the linear function becomes more gradual) with escalating $a$ values. After our analyses, we settled on $a=1.2$ since it amplifies the metric value when both distributions are identical and avoids saturation near $u=0$. The choice of $k$ doesn't profoundly sway the metric's sensitivity, which is also illustrated in Fig.~3 of the main paper. Therefore, we opted for $k=4$, a value intermediate between the choices for P\&R and D\&C.

\subsection{Outlier experiment with real-world distribution}
We further investigate the robustness of metrics by using CIFAR-10 \cite{krizhevsky2012imagenet} and DeepInversion (DI) model~\cite{yin2020dreaming} trained for CIFAR-10. DI is an inversion model trained to generate images that match the pre-trained neural network's intermediate feature statistics without explicitly observing the source images. Therefore, DI generates images that are somewhat distorted and unrecognizable (See Fig.~\ref{fig:append10}), yet preserves the statistics of CIFAR-10. We split generated images from DI into two subsets: inliers and outliers. We identify the outliers by computing the average of the $k_{th}$ nearest distance among real samples and selecting generated images whose nearest distance from real samples is greater than the average. Then, we substitute the real samples with these outliers. In Fig.~\ref{fig:append9}, we present the metric increments with respect to the number of outliers. We compare the increment of each metric with respect to the case where there are no outliers. The result shows that I-precision increases by about 0.9 with just one outlier, indicating its vulnerability to outliers. In addition, with only five outliers, Density shows a significant increase and eventually exceeds the increment of I-precision. This is because DSR accumulates the constant-density of hyperspheres, as we discussed in the main paper. Furthermore, we observe that I-precision and Density suddenly decrease as the number of outliers exceeds their hyperparameter $k$. This is not desirable behavior for a reliable and consistent metric. The sudden drop occurs because the size of the $k$NN hypersphere can change dramatically as the number of outliers exceeds $k$, highlighting the susceptible property of instance-specific $k$NN. On the other hand, P-precision is less affected by the outlier and increases linearly, demonstrating robustness to both outliers and $k$.

\subsection{Evaluating generative models}
Here, we provide additional quantitative results of the state-of-the-art generative models trained on various datasets using existing metrics. We use officially pre-trained models from their official codes\footnote{https://github.com/openai/guided-diffusion}\footnote{https://github.com/NVlabs/stylegan3} and take FID scores from their papers. The result of FID, IP\&IR, D\&C, and our PP\&PR is reported in Tab.~\ref{tab:append2}. For all metrics except FID, we measure the metrics between 50K generated samples and all available real samples, but up to 10K, as recommended in \cite{dhariwal2021diffusion}.

\input{Table/table_eval_gen_append.tex}

\begin{figure}[t]
\centering
\subfloat[Highest $L$]{\includegraphics[width=0.8\textwidth]{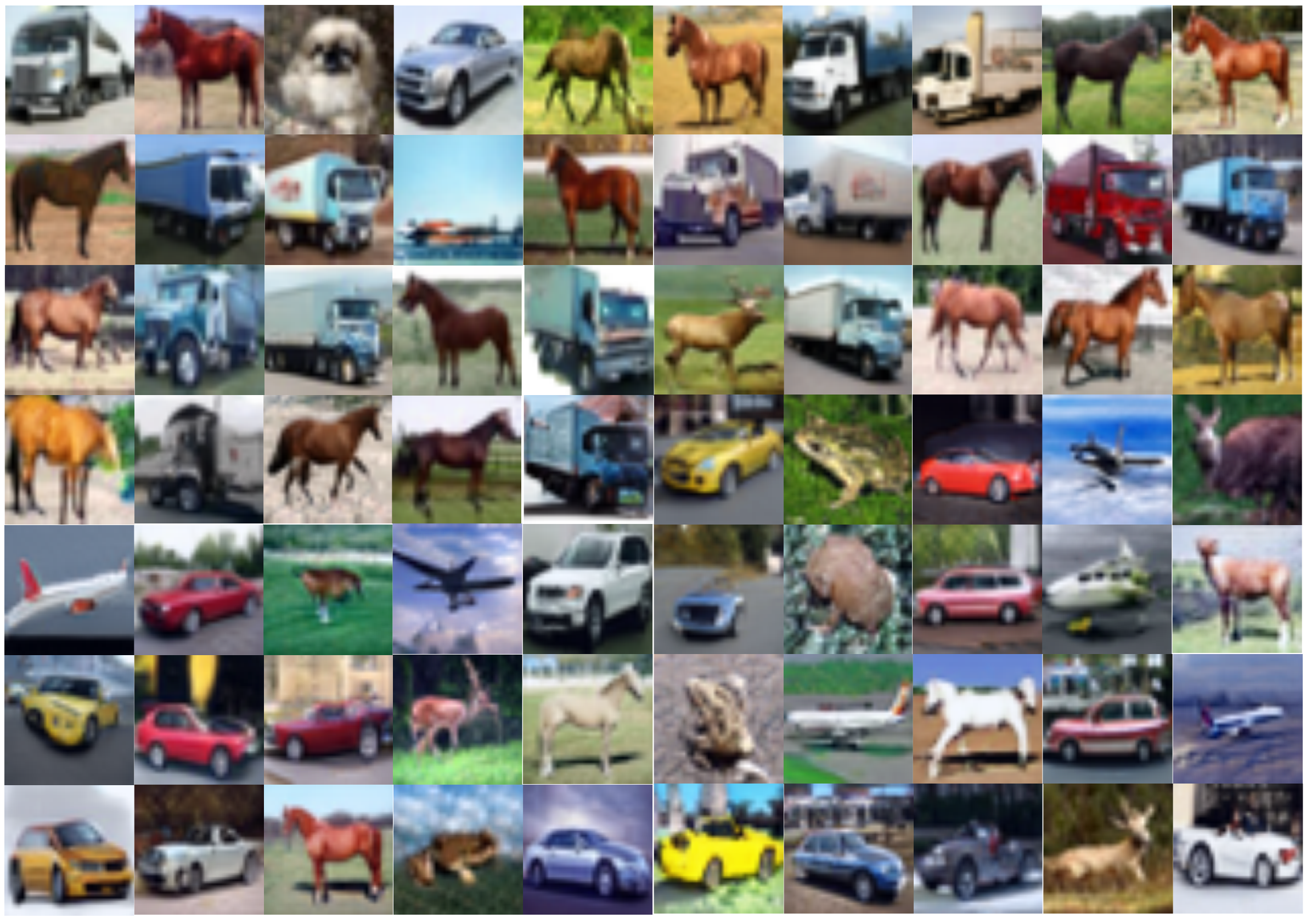}
}
\\
\subfloat[Lowest $L$]{\includegraphics[width=0.8\textwidth]{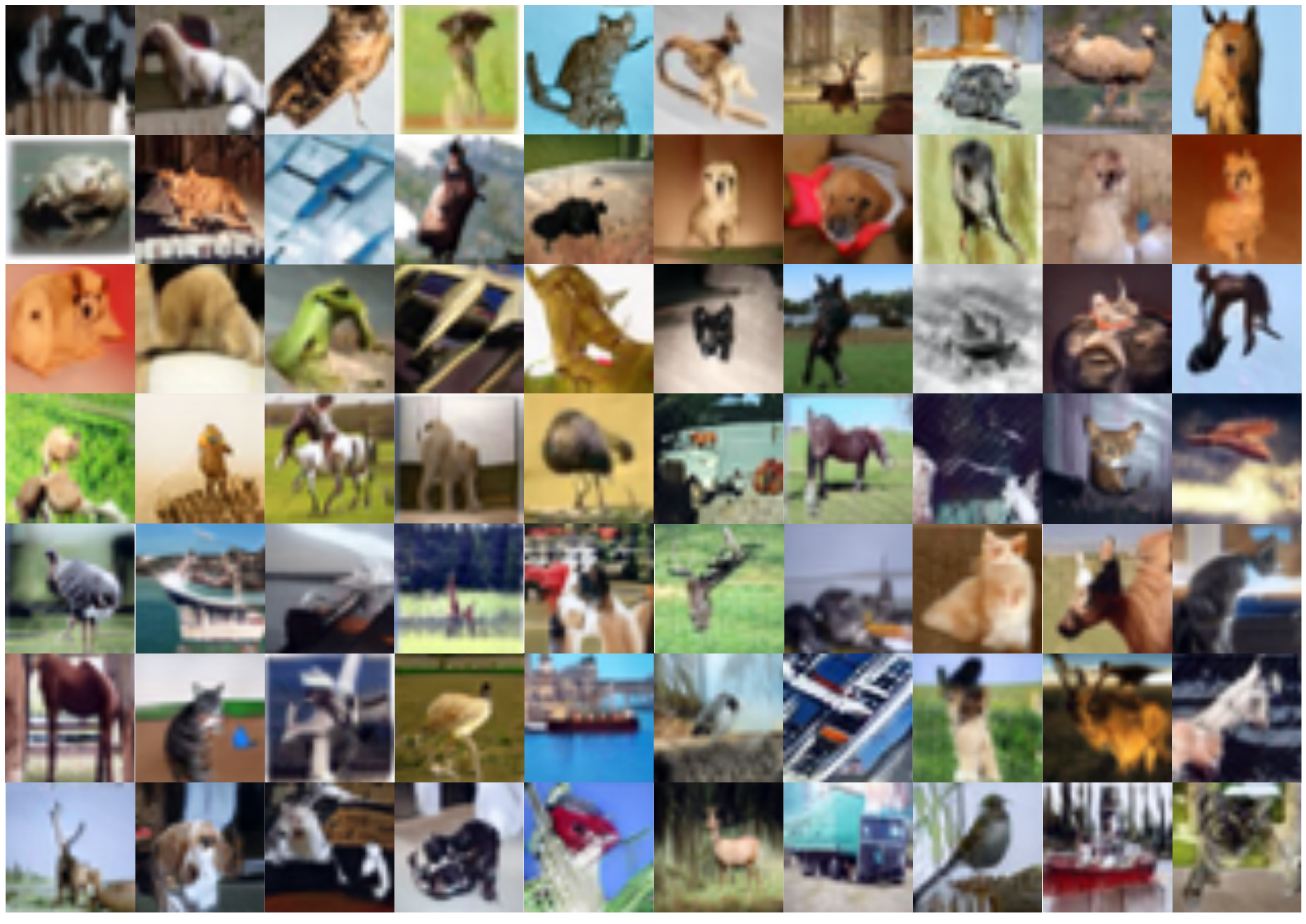}
}
\caption{Qualitative examples of BigGAN~\cite{brock2018large} generated images on CIFAR-10~\cite{krizhevsky2012imagenet} sorted according to $L$ (PSR - DSR). (a) are images with highest $L$ and (b) are images with lowest $L$.}
\label{fig:append11}
\end{figure}

\begin{figure*}[t]
\centering
\subfloat[Highest $L$]{\includegraphics[width=0.75\textwidth]{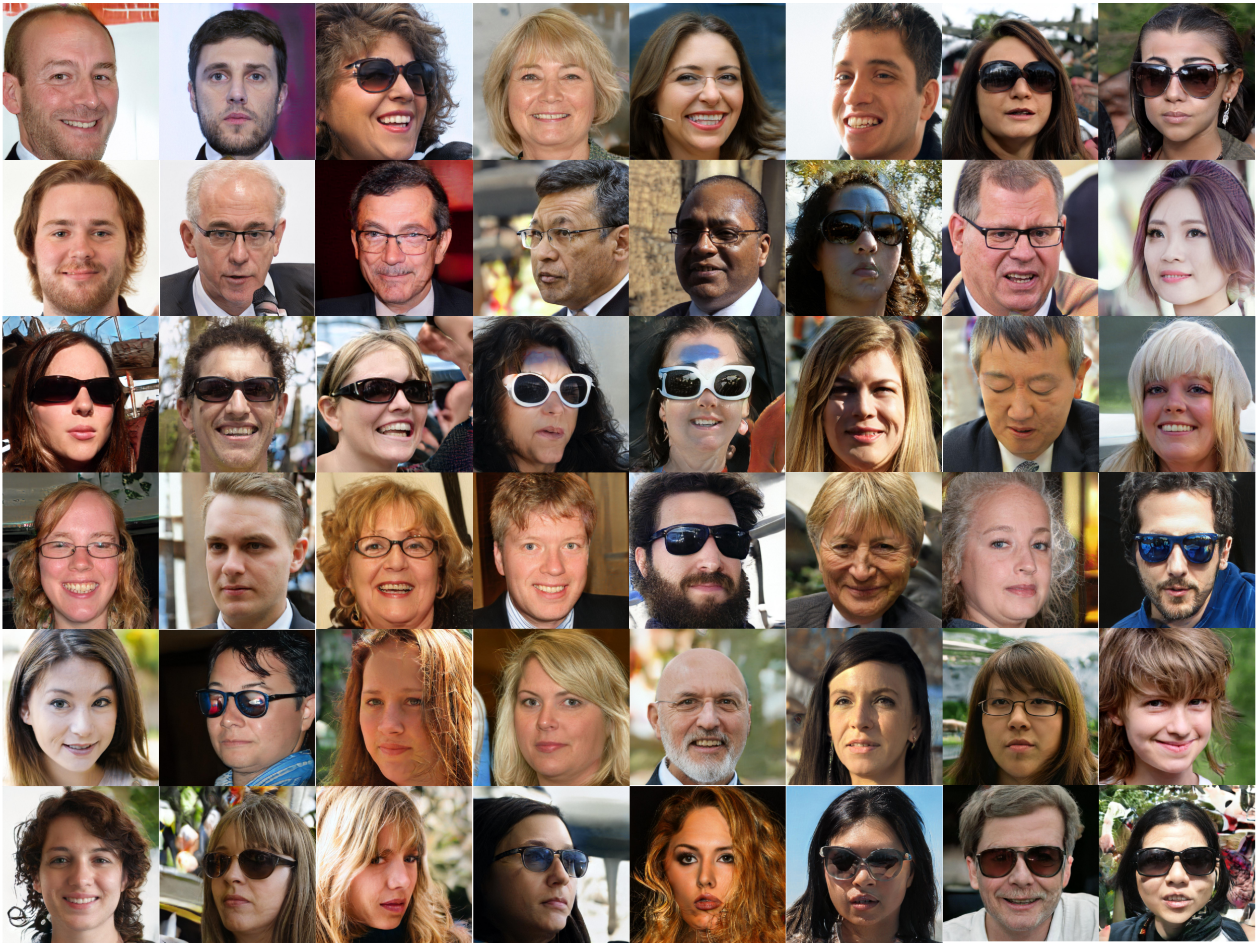}
}
\\
\subfloat[Lowest $L$]{\includegraphics[width=0.75\textwidth]{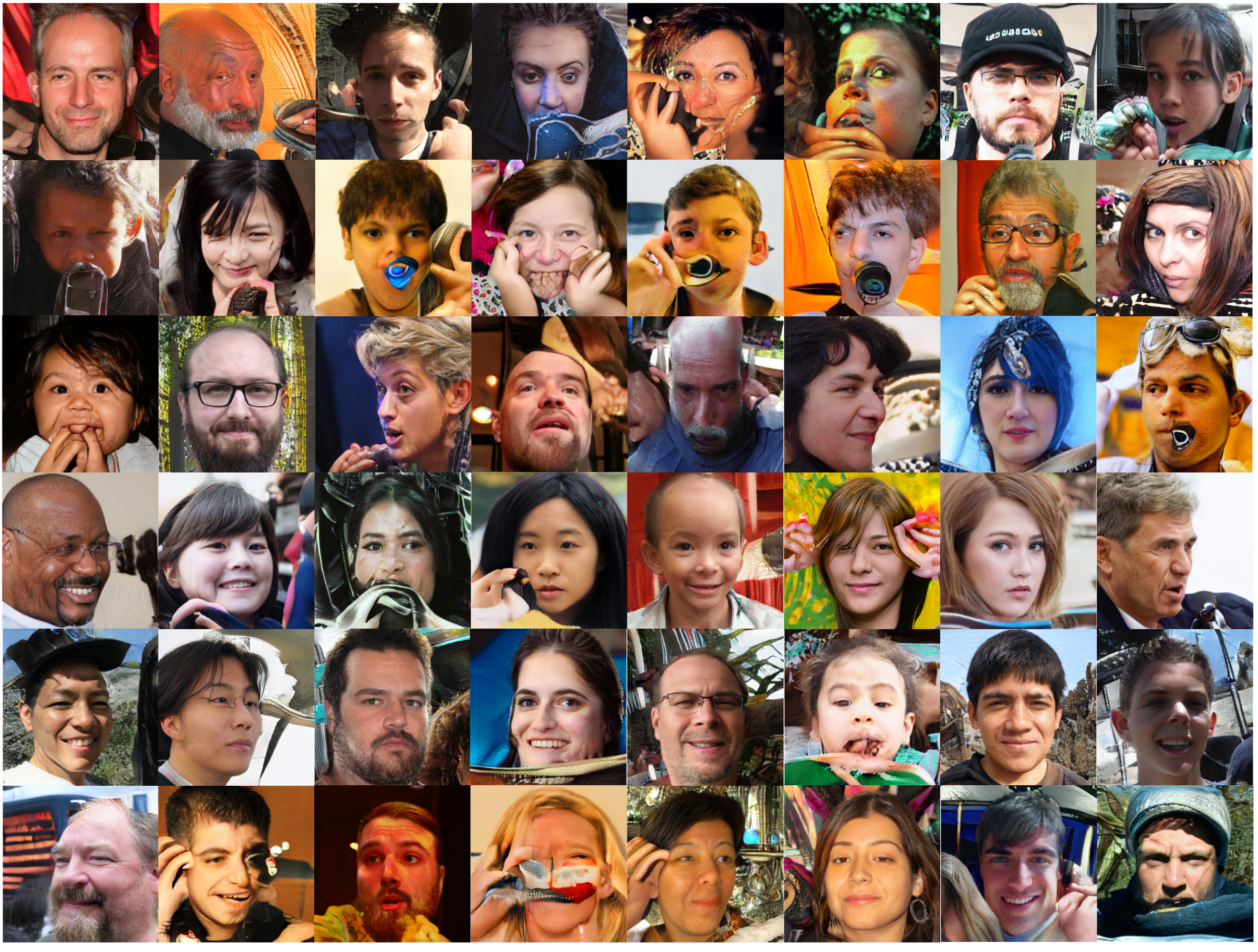}
}
\caption{Qualitative examples of StyleGAN~\cite{karras2019style} generated images on FFHQ~\cite{karras2019style} sorted according to $L$ (PSR - DSR). (a) are images with highest $L$ and (b) are images with lowest $L$.}
\label{fig:append12}
\end{figure*}

%% file: Table/table_userstudy.tex
\begin{table}[t]
\caption{The participants were asked to choose between two generated images that were sorted by $L$. Specifically, $L_{low}$ consists of images with high DSR value but low PSR value, while $L_{high}$ consists of images with high PSR value but low DSR.}
\centering
\resizebox{0.3\columnwidth}{!}{%
\begin{tabular}{ccc} 
\hline
                 & \multicolumn{2}{c}{PSR vs. DSR}  \\ 
\cline{2-3}
                 & $L_{low}$   & $L_{high}$                \\ 
\cline{2-3}
Preference Score $\uparrow$ & 17.40\% & \textbf{82.60}\%               \\
\hline
\end{tabular}
}\label{tab:append1}
\end{table}

%% file: Table/table_eval_gen_append.tex
\begin{table*}[t]
\caption{Quantitative result of various generative models on real-world datasets. The reported values are obtained by measuring each metric five times and taking the average. Bolded values indicates the top scores evaluated by the corresponding metric.}
\centering
\resizebox{0.99\textwidth}{!}{%
\begin{tabular}{l|c|ccc|ccc|ccc} 
\hline
\multicolumn{1}{l}{Model} & \multicolumn{1}{c}{FID$\downarrow$} & PP$\uparrow$                   & PR$\uparrow$                   & \multicolumn{1}{c}{$\text{F}_1$$\uparrow$} & IP$\uparrow$                   & IR$\uparrow$                   & \multicolumn{1}{c}{$\text{F}_1$$\uparrow$} & D$\uparrow$                    & C$\uparrow$                    & $\text{F}_1$$\uparrow$              \\
\multicolumn{11}{l}{\textbf{ImageNet 256x256}}                                                                                                                                                                                                                                   \\ 
\hline
\hline
ADM~\cite{dhariwal2021diffusion}                & 4.95                   & 0.538 (1.0e-5)                & \textbf{0.732} (1.2e-5)               & 0.621                       & 0.681 (3.4e-5)      & \textbf{0.688} (2.5e-4)                & 0.684                        & 1.52 (6.1e-3)      & 0.876 (1.1e-5)      & 1.112       \\
ADM-G~\cite{dhariwal2021diffusion}               & \textbf{4.58}          & 0.699 (3.1e-5)               & 0.587 (1.4e-5)      & \textbf{0.638}               & 0.818 (7.4e-5)               & 0.606 (1.7e-5)      & \textbf{0.696}               & 2.071 (1.4e-3)               & \textbf{0.956} (2.4e-5)              & 1.307                 \\
BigGAN~\cite{brock2018large}               & 8.12                   & \textbf{0.751} (3.4e-5)       & 0.465 (2.8e-5)                & 0.574                        & \textbf{0.874} (5.4e-5)               & 0.403 (2.1e-5)               & 0.551                        & \textbf{2.481} (3.4e-3)               & 0.945 (2.9e-5)               & \textbf{1.368}                \\
\multicolumn{1}{l}{}      & \multicolumn{1}{l}{}    & \multicolumn{1}{l}{} & \multicolumn{1}{l}{} & \multicolumn{1}{l}{}         & \multicolumn{1}{l}{} & \multicolumn{1}{l}{} & \multicolumn{1}{l}{}         & \multicolumn{1}{l}{} & \multicolumn{1}{l}{} & \multicolumn{1}{l}{}  \\
\multicolumn{11}{l}{\textbf{AFHQv2 512x512}}                                                                                                                                                                                                                                        \\ 
\hline
\hline
StyleGAN2~\cite{karras2020analyzing}                 & 4.62                    & \textbf{0.568} (1.6e-5)       & 0.689 (3.6e-5)               & 0.623                        & \textbf{0.716} (1.4e-4)      & 0.494 (3.1e-5)                & 0.584                        & \textbf{1.886} (4.4e-3)      & 0.790 (1.9e-5)               & \textbf{1.113}        \\
StyleGAN3-R~\cite{karras2021alias}               & 4.40                    & 0.564 (9.4e-6)               & 0.725 (1.4e-5)               & 0.634                        & 0.685 (6.4e-5)               & \textbf{0.591} (3.7e-5)      & \textbf{0.635}               & 1.576 (5.1e-4)               & 0.770 (7.1e-5)                & 1.034                 \\
StyleGAN3-T~\cite{karras2021alias}               & \textbf{4.04}           & 0.567 (6.2e-5)               & \textbf{0.727} (5.1e-5)      & \textbf{0.637}               & 0.699 (1.6e-4)               & 0.578 (8.1e-5)               & 0.632                        & 1.624 (1.1e-3)               & \textbf{0.792} (1.2e-5)      & 1.065 \\
\multicolumn{1}{l}{}      & \multicolumn{1}{l}{}    & \multicolumn{1}{l}{} & \multicolumn{1}{l}{} & \multicolumn{1}{l}{}         & \multicolumn{1}{l}{} & \multicolumn{1}{l}{} & \multicolumn{1}{l}{}         & \multicolumn{1}{l}{} & \multicolumn{1}{l}{} & \multicolumn{1}{l}{}  \\
\multicolumn{11}{l}{\textbf{LSUN Bedroom 256x256}}                                                                                                                                                                                                                                        \\ 
\hline
\hline
DDPM~\cite{karras2020analyzing}                 & 4.88                    & 0.799 (4.8e-5)      & 0.749 (1.4e-5)               & 0.773                        & 0.606 (2.2e-4)      & 0.444 (1.4e-4)               & 0.512                        & 1.701 (3.4e-3)      & 0.977 (6.4e-5)               & 1.241        \\
ADM~\cite{dhariwal2021diffusion}               & \textbf{1.91}                    & \textbf{0.839} (8.8e-5)               & 0.731 (6.4e-5)               & \textbf{0.781}                        & \textbf{0.659} (5.8e-5)               & 0.494 (3.4e-5)      & \textbf{0.565}               &\textbf{1.929} (2.8e-3)               & \textbf{0.993} (1.2e-5)               & \textbf{1.311}                 \\
StyleGAN2~\cite{karras2020analyzing}               & 2.35           & 0.801 (2.1e-5)                & \textbf{0.753} (1.8e-5)      & 0.776               & 0.591 (1.9e-5)               & \textbf{0.501} (2.7e-5)               & 0.543                        & 1.732 (8.6e-4)               & 0.986 (2.9e-5)      & 1.257 \\
\multicolumn{1}{l}{}      & \multicolumn{1}{l}{}    & \multicolumn{1}{l}{} & \multicolumn{1}{l}{} & \multicolumn{1}{l}{}         & \multicolumn{1}{l}{} & \multicolumn{1}{l}{} & \multicolumn{1}{l}{}         & \multicolumn{1}{l}{} & \multicolumn{1}{l}{} & \multicolumn{1}{l}{} \\
\multicolumn{11}{l}{\textbf{MetaFaces 1024x1024}}                                                                                                                                                                                                                                   \\ 
\hline
\hline
StyleGAN2~\cite{karras2020analyzing}                & 15.22                   & 0.896 (7.8e-5)               & 0.703 (6.7e-5)               & 0.788                        & \textbf{0.797} (4.4e-4)      & 0.291 (1.4e-4)               & 0.426                        & \textbf{2.171} (3.8e-3)      & 0.996 (1.1e-5)      & \textbf{1.366}        \\
StyleGAN3-R~\cite{karras2021alias}               & \textbf{15.11}          & 0.890 (3.6e-5)               & \textbf{0.770} (2.9e-5)      & \textbf{0.825}               & 0.738 (1.4e-4)               & 0.461 (1.6e-5)      & 0.567               & 1.864 (2.3e-3)               & 0.992 (3.4e-5)               & 1.295                 \\
StyleGAN3-T~\cite{karras2021alias}               & 15.33                   & \textbf{0.901} (3.4e-5)      & 0.747 (2.5e-5)               & 0.817                        & 0.748 (8.1e-5)               & \textbf{0.478} (5.4e-5)               & \textbf{0.583}                        & 1.885 (9.7e-4)               & \textbf{0.997} (1.1e-5)               & 1.304                 \\
\multicolumn{1}{l}{}      & \multicolumn{1}{l}{}    & \multicolumn{1}{l}{} & \multicolumn{1}{l}{} & \multicolumn{1}{l}{}         & \multicolumn{1}{l}{} & \multicolumn{1}{l}{} & \multicolumn{1}{l}{}         & \multicolumn{1}{l}{} & \multicolumn{1}{l}{} & \multicolumn{1}{l}{}  \\
\multicolumn{11}{l}{\textbf{ImageNet 64x64}}                                                                                                                                                                                                                                        \\ 
\hline
\hline
ADM~\cite{dhariwal2021diffusion}                       & \textbf{2.60}           & 0.738 (5.1e-5)               & \textbf{0.734} (3.4e-5)      & \textbf{0.736}               & 0.734 (1.1e-5)               & \textbf{0.647} (3.4e-5)       & \textbf{0.688}                & 1.867 (2.6e-3)                & \textbf{0.956} (5.4e-5)      & 1.265                 \\
BigGAN~\cite{brock2018large}                    & 4.07                    & \textbf{0.776} (5.4e-5)      & 0.612 (7.2e-5)               & 0.684                        & \textbf{0.792} (8.9e-5)      & 0.531 (2.1e-5)               & 0.635                        & \textbf{2.360} (4.4e-3)      & 0.954 (1.4e-5)               & \textbf{1.359}        \\
\multicolumn{1}{l}{}      & \multicolumn{1}{l}{}    & \multicolumn{1}{l}{} & \multicolumn{1}{l}{} & \multicolumn{1}{l}{}         & \multicolumn{1}{l}{} & \multicolumn{1}{l}{} & \multicolumn{1}{l}{}         & \multicolumn{1}{l}{} & \multicolumn{1}{l}{} & \multicolumn{1}{l}{}  \\
\multicolumn{11}{l}{\textbf{ImageNet 128x128}}                                                                                                                                                                                                                                      \\ 
\hline
\hline
ADM~\cite{dhariwal2021diffusion}                       & \textbf{5.91}           & 0.601 (5.8e-5)               & \textbf{0.737} (4.9e-5)      & \textbf{0.662}               & 0.700 (8.4e-5)               & \textbf{0.697} (7.2e-5)      & \textbf{0.699}               & 1.613 (9.1e-4)               & 0.925 (3.4e-5)               & 1.176                 \\
BigGAN~\cite{brock2018large}                    & 6.01                    & \textbf{0.772} (1.1e-5)      & 0.473 (2.3e-5)               & 0.586                        & \textbf{0.862} (3.6e-5)      & 0.452 (1.6e-5)               & 0.593                        & \textbf{2.521} (1.7e-3)      & \textbf{0.954} (3.8e-5)      & \textbf{1.384}        \\
\multicolumn{1}{l}{}      & \multicolumn{1}{l}{}    & \multicolumn{1}{l}{} & \multicolumn{1}{l}{} & \multicolumn{1}{l}{}         & \multicolumn{1}{l}{} & \multicolumn{1}{l}{} & \multicolumn{1}{l}{}         & \multicolumn{1}{l}{} & \multicolumn{1}{l}{} & \multicolumn{1}{l}{}  \\     
\end{tabular}
}\label{tab:append2}
\end{table*}